\documentclass{article}
\usepackage{authblk}
\usepackage[colorlinks,citecolor=green]{hyperref}
\usepackage{amsmath}
\usepackage{picins}
\usepackage{array}
\usepackage{makecell}
\usepackage{booktabs}
\usepackage{mathrsfs}
\usepackage{graphicx}
\usepackage{multirow}
\usepackage{subfigure}
\usepackage{amsfonts}
\usepackage{cleveref}
\usepackage{algorithm}
\usepackage{algorithmic}
\crefname{equation}{Eq.}{Eqs.}
\crefname{section}{Section}{Sections}
\crefname{algorithm}{Algorithm}{Algorithms}
\crefname{figure}{Fig.}{Figs.}
\crefname{table}{Table}{Tables}
\usepackage{graphicx}
\usepackage{caption}

\begin{document}


\title{Object-Adaptive LSTM Network for Real-time Visual Tracking with Adversarial Data Augmentation}

\author[1,2]{Yihan Du}
\author[1]{Yan Yan \thanks{Corresponding author: yanyan@xmu.edu.cn}}
\author[3]{Si Chen}
\author[4]{Yang Hua}

\affil[1]{School of Informatics, Xiamen University, Fujian 361005, China}
\affil[2]{Institute for Interdisciplinary Information Sciences, Tsinghua University, Beijing 100084, China}
\affil[3]{School of Computer and Information Engineering, Xiamen University of Technology, Fujian 361024, China}
\affil[4]{School of Electronics, Electrical Engineering and Computer Science, Queen's University Belfast, UK}


\maketitle

\begin{abstract}
In recent years, deep learning based visual tracking methods have obtained great success owing to the powerful feature representation ability of Convolutional Neural Networks (CNNs). Among these methods, classification-based tracking methods exhibit excellent performance while their speeds are heavily limited by the expensive computation for massive proposal feature extraction. In contrast, matching-based tracking methods (such as Siamese networks) possess remarkable speed superiority. However, the absence of online updating renders these methods unadaptable to significant object appearance variations. In this paper, we propose a novel real-time visual tracking method, which adopts an object-adaptive LSTM network to effectively capture the video sequential dependencies and adaptively learn the object appearance variations. For high computational efficiency, we also present a fast proposal selection strategy, which utilizes the matching-based tracking method to pre-estimate dense proposals and selects high-quality ones to feed to the LSTM network for classification. This strategy efficiently filters out some irrelevant proposals and avoids the redundant computation for  feature extraction, which enables our method to operate faster than conventional classification-based tracking methods. In addition, to handle the problems of sample inadequacy and class imbalance during online tracking, we adopt a data augmentation technique based on the Generative Adversarial Network (GAN) to facilitate the training of the LSTM network. Extensive experiments on four visual tracking benchmarks demonstrate the state-of-the-art performance of our method in terms of both tracking accuracy and speed, which exhibits great potentials of recurrent structures for visual tracking.
\end{abstract}



\section{Introduction}
Visual tracking aims to track an arbitrary object throughout a video sequence, where the target is solely identified by the annotation in the first frame. As a fundamental problem in computer vision, visual tracking has extensive applications such as video surveillance, human-computer interaction and automation. Despite rapid progress in the past few decades, visual tracking is still very challenging since the trackers are prone to show inferior performance under complex scenes including occlusion, deformation, background clutter, \emph{etc}.

In recent years, deep learning has brought a significant breakthrough in tracking accuracy owing to the powerful feature representation ability of Convolutional Neural Networks (CNNs) \cite{cnn1998}. The deep tracking methods \cite{mdnet,sanet,goturn,siamesefc} can be roughly divided into two categories, \emph{i.e.}, classification-based tracking methods and matching-based tracking methods. Classification-based tracking methods \cite{mdnet,sanet,vital} train an online classifier to distinguish the object from the background. However, most of these methods contain complex feature extraction stages for massive proposals and sophisticated online updating techniques to adapt the network to the arbitrary temporally changing object. As a result, although these methods have achieved promising accuracy, the heavy computational burden renders these methods difficult to satisfy the real-time requirement of the tracking task. In addition, some high-accuracy trackers \cite{mdnet,sanet,vital} pre-train their networks based on the videos from the visual tracking benchmarks, which may raise the risk of over-fitting.

\begin{figure}[!tb]
	\centering
	\includegraphics[width=0.8\textwidth]{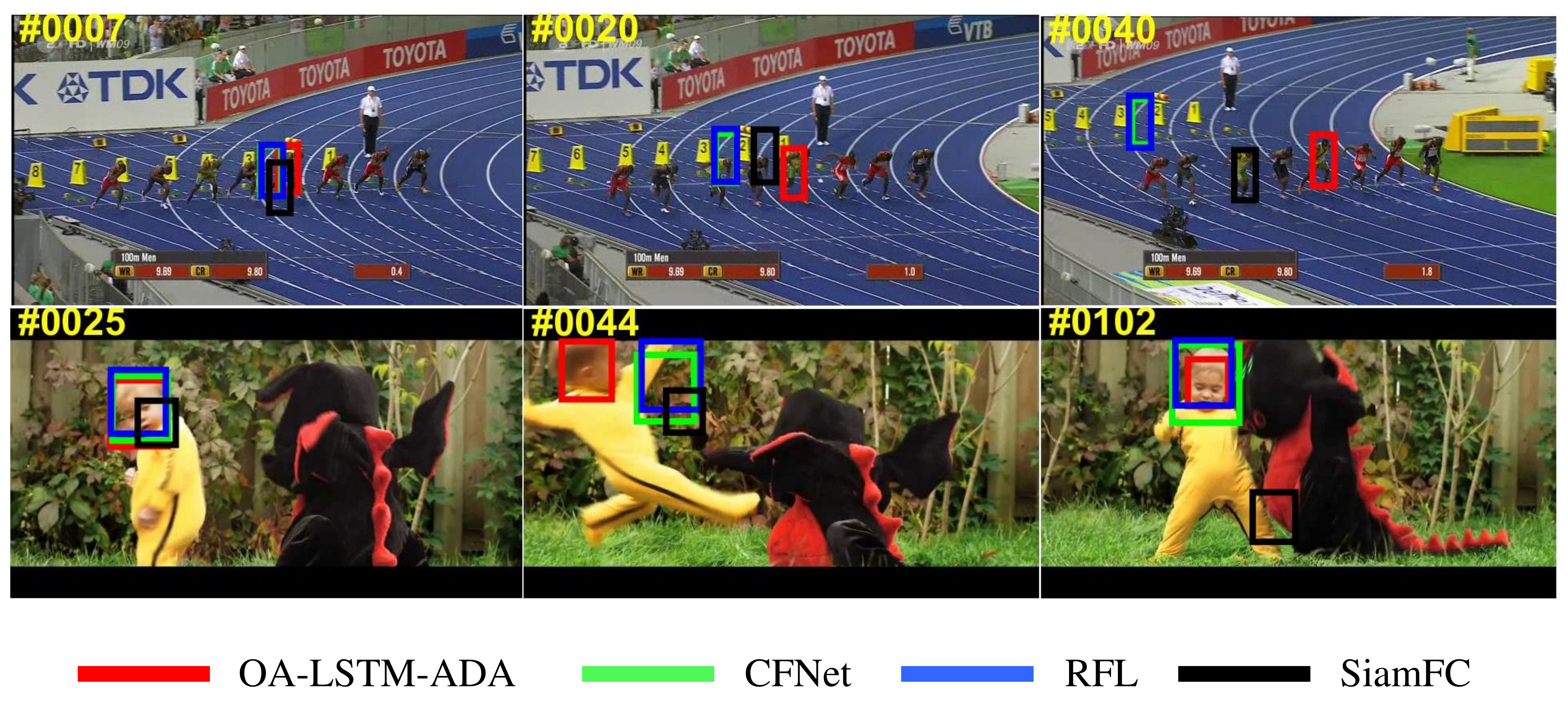}
	\caption{Comparison between our method (OA-LSTM-ADA) and the state-of-the-art matching-based tracking methods, \emph{i.e.}, CFNet \cite{cfnet}, RFL \cite{Recurrent_filter} and SiamFC \cite{siamesefc},  on the Bolt and DragonBaby \cite{otb15}  sequences. Our tracker that utilizes background information with online adaptability  performs more robustly than the other trackers when encountering object deformation and background clutter.}
	\label{comparison_siamfc}
\end{figure}

Matching-based tracking methods \cite{goturn,siamesefc, sint} usually firstly learn general matching models offline on the large dataset (such as ILSVRC15 \cite{imagenet}). Then, these methods directly match the candidate proposals with the target template using the pre-trained models during online tracking. The succinct online tracking algorithms make these methods possess remarkable speed superiority. However, due to the inherent lack of online adaptability and the ignorance of background information, these matching-based tracking methods cannot well handle the object appearance variations and similar objects in the background. Thus, these methods usually suffer from drift when the object appearance changes or the similar object appears in some complex scenes. Recent matching-based tracking methods \cite{cfnet,Recurrent_filter} are proposed to online update the matching template of the object, but they still do not  utilize the background information sufficiently. \cref{comparison_siamfc} shows a  comparison between our method and some state-of-the-art matching-based tracking methods, \emph{i.e.}, CFNet \cite{cfnet}, RFL \cite{Recurrent_filter} and SiamFC \cite{siamesefc}. The compared matching-based tracking methods cannot effectively track the target when encountering the significant object appearance variations or complex background, while our method can accurately locate the target position in these challenging situations.

Most of existing deep learning based tracking methods take advantage of the powerfulness of CNN in feature representation, while these methods cannot fully utilize the temporal dependencies among successive frames in a video sequence. Different from the traditional CNN-based tracking methods, we consider the Long Short-Term Memory (LSTM) \cite{lstm1997} network, a variant of the Recurrent Neural Network (RNN) \cite{rnn}, which can memorize useful historical information and capture long-range sequential dependencies. Based on the LSTM network, we are able to utilize the sequential dependencies and learn the target appearance variations via maintaining an internal object representation model.

In this paper, we propose a novel object-adaptive LSTM network for visual tracking, which can fully utilize the time dependencies among successive frames of a video sequence and effectively adapt to the temporally changing object via memorizing the target appearance variations. Since the proposed LSTM network is learned online \footnote{In this paper, ``online'' refers to that only the information accumulated up to the present frame
is used for inference during tracking.} as a per-object classifier, our tracker can effectively track an arbitrary object with superior adaptability to sequence-specific circumstances. Furthermore, due to its intrinsic recurrent structure, our network can dynamically update the internal state, which characterizes the object representation during the forward passes. For high computational efficiency, we also present a fast proposal selection strategy. In particular, we make use of the matching-based tracking method to pre-estimate the dense proposals and select high-quality ones to feed to the LSTM network for classification. In this strategy, we directly obtain the proposal features from the big feature map of the search region so that only one feature extraction operation is performed. In this way, the proposed strategy can effectively filter out the irrelevant proposals and only retain the high-quality ones. As a result, the computational burden of proposal feature extraction is largely alleviated.

\begin{figure*}[!t]
	\centering
	\includegraphics[width=1.0\textwidth]{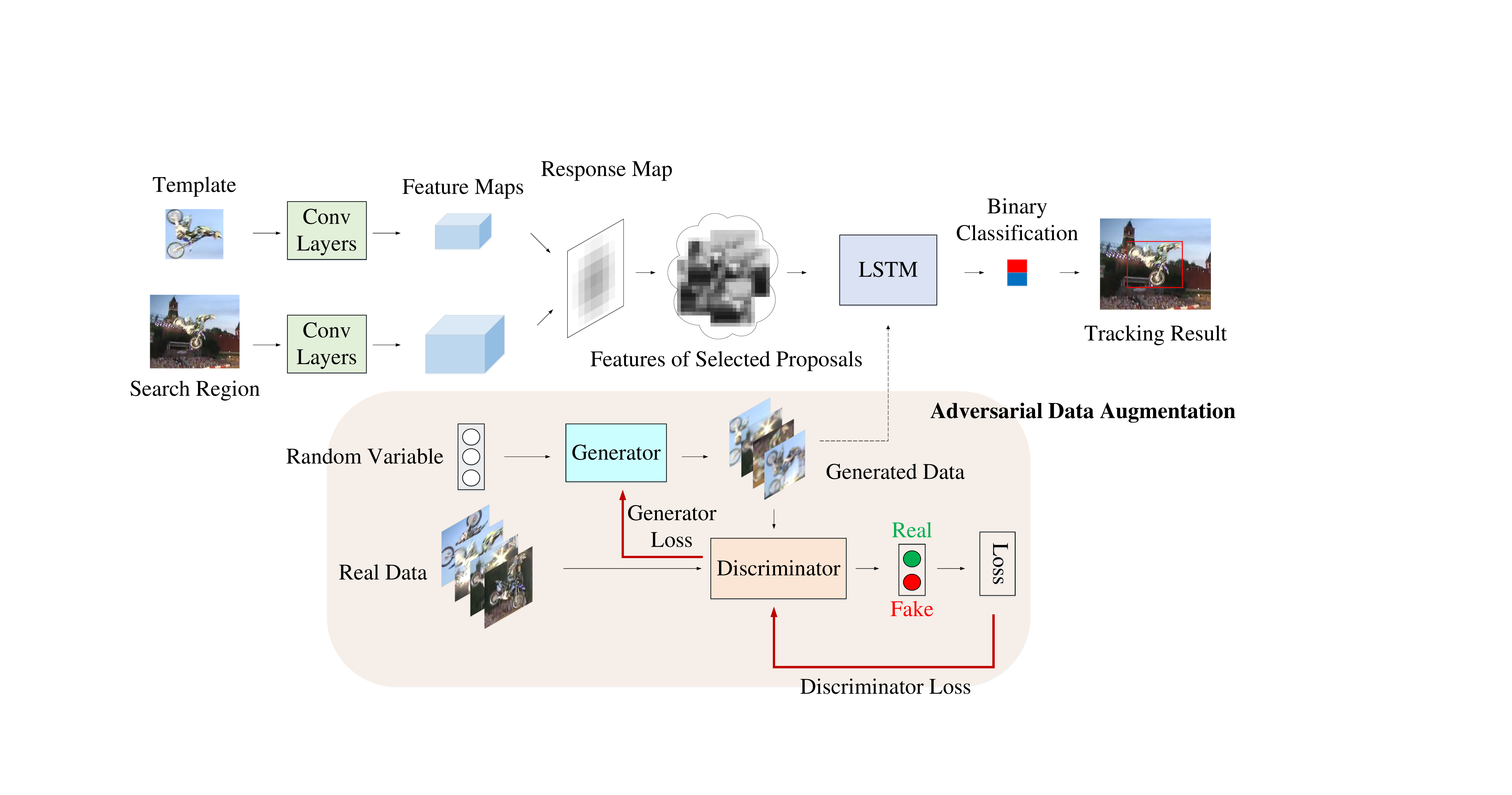}
	\caption{Pipeline of the proposed method for visual object tracking. During online tracking, we maintain a set of high-confident tracking results including the given original object. The real data fed to the discriminator are drawn according to this tracking result set. The ``Loss'' at the far right of  the ``Adversarial Data Augmentation'' part collectively refers to the discriminator loss and generator loss of GAN. The black solid arrows represent the links between blocks. The black dashed arrow between ``Generated Data'' and ``LSTM'' means that the generated data of GAN augment the training samples of LSTM. The red solid arrows stand for the backpropagation direction of losses during the training of GAN.}
	\label{pipeline}
\end{figure*}

In order to handle the sample inadequacy and class imbalance problems during the online learning process, we also adopt Generative Adversarial Network (GAN) \cite{gan2014} to generate diverse positive samples, which augments the available training data and thus facilitates the training of the LSTM network. In this paper, GAN is trained in the first frame and updated in the subsequent frames during tracking.
We refer to our method as an Object-Adaptive LSTM network with Adversarial Data Augmentation (OA-LSTM-ADA) for visual tracking. \cref{pipeline} illustrates the pipeline of our tracking method. Experimental results on the OTB (both OTB-2013 and OTB-2015) \cite{otb15}, TC-128 \cite{tc128}, UAV-123 \cite{uav123} and VOT-2017 \cite{vot2017} benchmarks demonstrate that our method achieves the state-of-the-art performance while operating at real-time speed, which exhibits great potentials of recurrent structures for visual object tracking.

We summarize our main contributions as follows:

\begin{itemize}
	\item
	We propose a novel object-adaptive LSTM network for visual tracking, which fully exploits the sequential dependencies and effectively adapts to the object appearance variations. Due to its intrinsic recurrent structure, the internal state of the network can be dynamically updated during the forward passes. Therefore, the proposed method is able to robustly track an arbitrary object under complex scenarios.
	
	\item
	We propose a fast proposal selection strategy, which utilizes the matching-based tracking method to pre-estimate the dense samples and selects  high-quality ones to feed to the LSTM network.  The proposed strategy directly obtains the proposal features from the feature map of search region. In this manner, the expensive computational cost for proposal feature extraction in conventional classification-based tracking frameworks is effectively reduced, by which our method can operate in real-time.

	\item
	We propose a data augmentation strategy to address the problems of sample inadequacy and class imbalance during online learning of the LSTM network. We use an online learned GAN to generate diverse positive samples with sequence-specific information, which enriches the available training data and thus facilitates the training of the LSTM network.
	
	
\end{itemize}

This paper is an extension of our previous work \cite{oa-lstm}.  In this paper, we accelerate the proposed method by directly obtaining the proposal features from the  feature map of the search region. No extra computational cost for proposal feature extraction is required. Thus,  our method can operate in real-time. Moreover, we additionally investigate the problems of sample inadequacy and class imbalance during the online training of the LSTM network. Specifically, we propose to use a GAN to augment the available training data, which significantly improves the performance of the original method.   The experiments are also extended via presenting results of the further internal comparison, state-of-the-art comparison and attribute-based comparison.

The rest of this paper is organized as follows: \cref{section_related work} gives an overview of the related work. \cref{section_proposed method} discusses the proposed tracking method, which contains the components of  the fast proposal selection strategy, the  object-adaptive LSTM network and the data augmentation technique. \cref{section_algorithm} describes the proposed online tracking algorithm. \cref{section_experiment} presents the experimental results on four public tracking benchmarks. Conclusions and future work are  drawn in \cref{section_conclusion}.

\section{Related Work}
\label{section_related work}
In this section, we briefly review the deep learning based tracking methods and discuss the related works on RNNs and generative adversarial learning.

\textbf{Visual tracking.} Visual tracking has been actively studied over the past few decades and it remains one of the most important and challenging problems in computer vision. A large number of visual tracking methods, including sparse representation \cite{Visual_tracking_using_strong_classifier,Linearization_to_nonlinear_learning,nonlinear_learning,Discriminative_tracking_using_tensor_pooling,under_motion_blur,Generalized_pooling}, multiple instance learning \cite{Neuro_multiple_instance_learning,Neuro_multiple_instance_learning2,Neuro_multiple_instance_learning3} and correlation filters \cite{dsst,kcf,staple,Neuro_correlation_filter}, have been proposed.
For example, a strong classifier and structural local sparse descriptors are introduced for tracking objects in \cite{Visual_tracking_using_strong_classifier}.
In \cite{nonlinear_learning}, a tracking method which jointly learns a nonlinear classier and a visual dictionary in the sparse coding manner, is proposed.
In \cite{Discriminative_tracking_using_tensor_pooling}, the authors use sparse coding tensors to represent target templates and candidates, and build the appearance model via incrementally learning.
A tracking framework  which combines blur state estimation and multi-task reverse sparse learning, is proposed in \cite{under_motion_blur}.
 A generalized feature pooling method \cite{Generalized_pooling} is presented for robust visual tracking.
 A novel two-stage classifier with the circulant structure \cite{Occlusion-aware} is developed to address scenes including occlusion.
In \cite{sampling_in_part_space}, the authors employ a part space with two online learned probabilities to represent the target structure.
 A hyperparameter optimization method \cite{Hyperparameter_optimization} is proposed for robust object tracking.


In recent years, deep learning based tracking methods \cite{mdnet,sanet,siamesefc,Manifold_Regularized_Convolutional_Neural_Networks} have shown their outstanding performance by taking advantage of the  powerful ability of CNNs in feature representation. These methods can be roughly divided into classification-based tracking methods and matching-based tracking methods. Classification-based tracking methods \cite{mdnet,sanet} treat visual tracking as a binary classification problem, which aims to distinguish the object from the background. For example, MDNet \cite{mdnet} adopts a multi-domain learning strategy to utilize large-scale annotated tracking data and learn an online per-object classifier. SANet \cite{sanet} proposes a structure-aware network to handle similar distractors.
MRCNN \cite{Manifold_Regularized_Convolutional_Neural_Networks} introduces a particle filter based tracking framework by taking advantage of an online updated manifold regularized deep model.
Although these methods achieve high tracking accuracy, the expensive cost spent on the massive proposal feature extraction and sophisticated online fine-tuning heavily limits their speeds. Besides, these methods perform the pre-training stages on tracking benchmark datasets, which may raise the risk of over-fitting.

Matching-based tracking methods \cite{siamesefc,goturn,sint} are developed to match the candidate proposals with the target template using the general pre-trained networks. These methods usually do not perform any online updating procedures so that they possess remarkable speed superiority. Siamese network is one of the most representative methods. For example, GOTURN \cite{goturn} uses the Siamese network to directly regress the object location from the previous frame. SiamFC \cite{siamesefc} proposes a fully-convolutional Siamese network to learn a general similarity function.
Despite the efficiency of these methods, the inherent lack of online adaptability makes them prone to drift when the object appearance significantly changes or similar objects appear.

Recently, several Siamese network based trackers \cite{dsiam,east,Siam-tri,SiameseRPN,Hierarchical_Attention_Siamese_Network,Quadruplet_network} have been proposed to address the above problems, which can improve the tracking accuracy while preserving real-time speeds. For example, DSiam \cite{dsiam} proposes a dynamic Siamese network with transformation learning and EAST \cite{east} learns a decision-making strategy in a reinforcement learning framework for adaptive tracking.
SiamFC-tri \cite{Siam-tri} incorporates a novel triplet loss  into the Siamese network  to extract expressive deep features.
SiameseRPN \cite{SiameseRPN} proposes an offline trained Siamese Region Proposal Network (RPN). DaSiameseRPN \cite{DaSiameseRPN} improves SiameseRPN by introducing a distractor-aware module. C-RPN \cite{Cascaded_SiameseRPN} proposes Siamese cascaded RPNs to solve the  problem of class imbalance by performing hard negative sampling.
HASiam \cite{Hierarchical_Attention_Siamese_Network} introduces the attention mechanism into the Siamese network to enhance its matching discrimination.
Quad \cite{Quadruplet_network} proposes a quadruplet network to detect the potential connections of training instances for better representation.
In contrast to the above Siamese based methods, we use the Siamese network to select high-quality proposals for computational efficiency and learn a real-time object-adaptive LSTM network to classify these selected proposals. As a result, the proposed tracker effectively captures the object appearance variations with online adaptability.

Recently, some works \cite{Deep_visual_attention_prediction,Video_salient_object_detection,attention_and_aesthetics_aware_photo_cropping} adopt specialized attention networks  for  saliency prediction. Different from these  works, we employ the fast proposal selection strategy for  salient object  detection, which efficiently  selects high-quality proposals and filters out the irrelevant ones according to the matching-based response map.

\textbf{Recurrent neural networks.} Recurrent Neural Networks (RNNs) have drawn extensive attention due to their excellent capability of memorizing useful historical information and modeling sequential data. Gan \emph{et al.} \cite{firststep} and Kahou \emph{et al.} \cite{ratm} use attention-based RNNs for visual tracking, but these methods only demonstrate their effectiveness on simple datasets (such as MNIST) instead of natural videos. Re3 \cite{re3} proposes a recurrent regression model to offline learn the changes in the target appearance and motion. SANet \cite{sanet} incorporates RNNs into CNNs to model the object structure and improve the tracking robustness. Note that RFL \cite{Recurrent_filter} and MemTrack \cite{memtrack} also combine Siamese networks and LSTM networks to track objects. They adopt pre-trained LSTM networks as target information memorizers to update the template-matching procedure in Siamese networks.  However, different from the above methods, in this paper we use Siamese network as a coarse object pre-estimator to filter out irrelevant proposals and train an LSTM network online as a fine object-specific classifer to distinguish the object from the background. Our LSTM classifier can not only sequence-specifically utilize both foreground and background information, but also effectively equip the proposed tracker with adaptability to the object appearance variations while operating in real-time.

\textbf{Generative adversarial learning.}  Recently, generative adversarial learning has been widely applied to visual tracking.
The state-of-the-art tracker, VITAL \cite{vital}, proposes to use GAN to identify the masks that maintain robust features of the object over a long temporal span. Although VITAL achieves high tracking accuracy, it is very slow due to massive feature extractions and sophisticated online fine-tuning procedures. SINT++ \cite{sint++} generates diverse positive samples via a deep generative model and learns a hard positive transformation network with reinforcement learning to occlude the object with background image patch for higher robustness. However, its slow basic tracker (\emph{i.e.}, SINT \cite{sint})  heavily limits its tracking speed, which is far from the real-time requirement.
In this paper, we directly employ GAN as an image data augmenter to generate diverse positive samples in the image space, while maintaining a real-time tracking speed. The generated realistic-looking sample images enrich the available training data and thus facilitate the training of the LSTM network.

\section{The Proposed Method}
\label{section_proposed method}
\subsection{Overview}

\begin{figure*}[!t]
	\centering
	\includegraphics[width=1\textwidth]{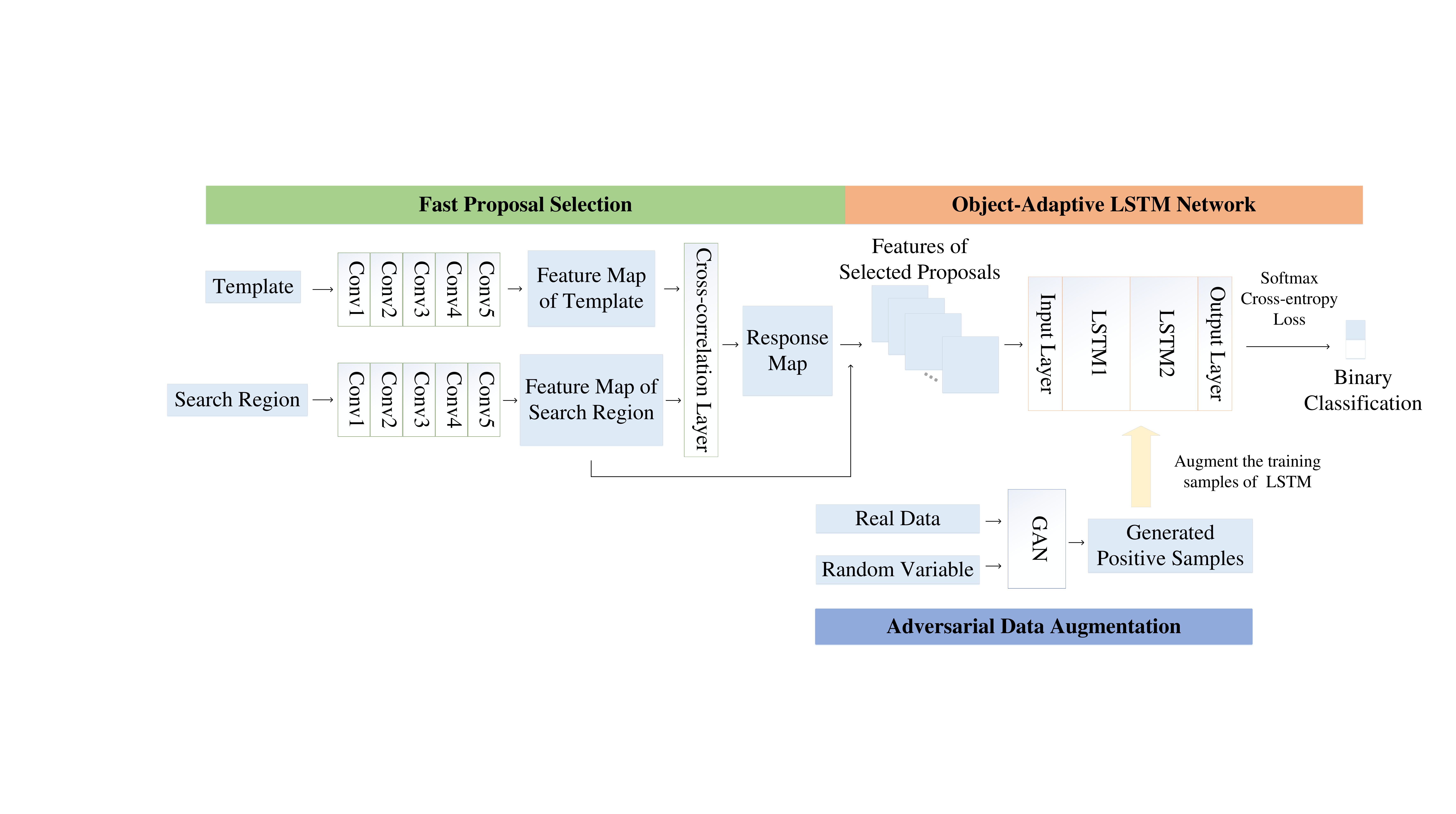}
	\caption{Overview of the proposed method.}
	\label{pipeline2}
\end{figure*}

As shown in \cref{pipeline2}, the proposed method consists of two stages, \emph{i.e.}, fast proposal selection via a pre-trained Siamese network and object classification via an online object-adaptive LSTM network.

In the first stage, we utilize the Siamese network to match the target template with the search region centered at the previously estimated target position. As a result, we can obtain a response map, which denotes the similarities between the target template and the proposals in the search region. Based on the response map, we select the high-quality proposals and crop their features from the big feature map of the search region to feed to the subsequent LSTM network for classification. This proposal selection strategy not only efficiently filters out the irrelevant proposals, but also significantly reduces the computational cost for proposal feature extraction. Therefore, our method can operate in real-time, which is faster than conventional classification-based tracking methods \cite{mdnet,sanet}.

In the second stage, we learn an object-adaptive LSTM network online to classify the input proposal features based on sequence-specific information. Taking advantage of the superior ability of LSTM to memorize useful historical information, we feed the LSTM network with the selected proposals, together with the previously estimated target state. By doing this, the LSTM network is able to identify the optimal target state according to the internal network state which effectively memorizes the object appearance variations over a long temporal span. Owing to the intrinsic recurrent structure of the LSTM network, the internal network state can be simultaneously updated when a forward pass is performed. Note that the Siamese network used in our method is pre-trained on a large dataset (\emph{i.e.}, ILSVRC15 \cite{imagenet}) and the proposed object-adaptive LSTM network is learned online. Therefore, our method is able to robustly track an arbitrary object without suffering from the problem of over-fitting to the tracking datasets.

In order to address the problems of sample inadequacy and class imbalance during the online learning process of  LSTM network, we make use of GAN to generate diverse positive samples to approximate the real target images. The generated diverse positive samples are incorporated into the training dataset of LSTM network. Such a strategy effectively augments the available training data and thus improves the tracking performance of our method.

\subsection{Fast Proposal Selection} \label{fast proposal}

In the conventional classification-based tracking framework (such as \cite{mdnet,sanet}), trackers usually generate massive candidate proposals via dense sampling and then evaluate these proposals through convolutional feature extractors and binary classifiers. However, the densely sampled proposals include many irrelevant and trivial proposals, which are far away from the object center. As a result, the unnecessary high computational cost is spent on the step of massive proposal feature extraction, which heavily constrains the tracking speed.

Recently, a number of matching-based tracking methods \cite{goturn,siamesefc,sint} are developed to directly compare the target template with the search region (and these methods usually do not involve online updating procedures). These methods possess remarkable speed superiority, but they lack of online adaptability to significant object appearance variations. Motivated by this observation, we utilize a representative matching-based tracking method, SiamFC \cite{siamesefc}, to pre-estimate the dense proposals and obtain their confidence scores. Then, we select the proposals of high confidence scores and crop their features from the big feature map of the search region to feed to the subsequent LSTM network for further classification.

Specifically, SiamFC \cite{siamesefc} trains a fully-convolutional Siamese network offline to compare the target template with the search region. By taking advantage of a bilinear layer which calculates the cross-correlation of inputs from two streams, it is able to achieve dense sliding-window evaluation in a single forward pass. The Siamese network can be formulated as the following similarity function,
\begin{align}
	\label{eq_siamfc}
	F(z,x)=\varphi(z)\ast \varphi(x)+k\mathbb{I},
\end{align}
where $z$ is a template image and $x$ is a search region. $\varphi$ refers to a convolutional embedding function and $F$ represents a similarity metric. `$\ast$' is the cross-correlation operation. $k\mathbb{I}$ denotes a signal that takes the value $k \in \mathbb{R}$ in every position. $F(z,x)$, denoting the output of the Siamese network, is a score map, which contains the similarities between the target template and each candidate proposal in the search region.

\begin{figure}[!tb]
	\centering
	\includegraphics[width=1\textwidth]{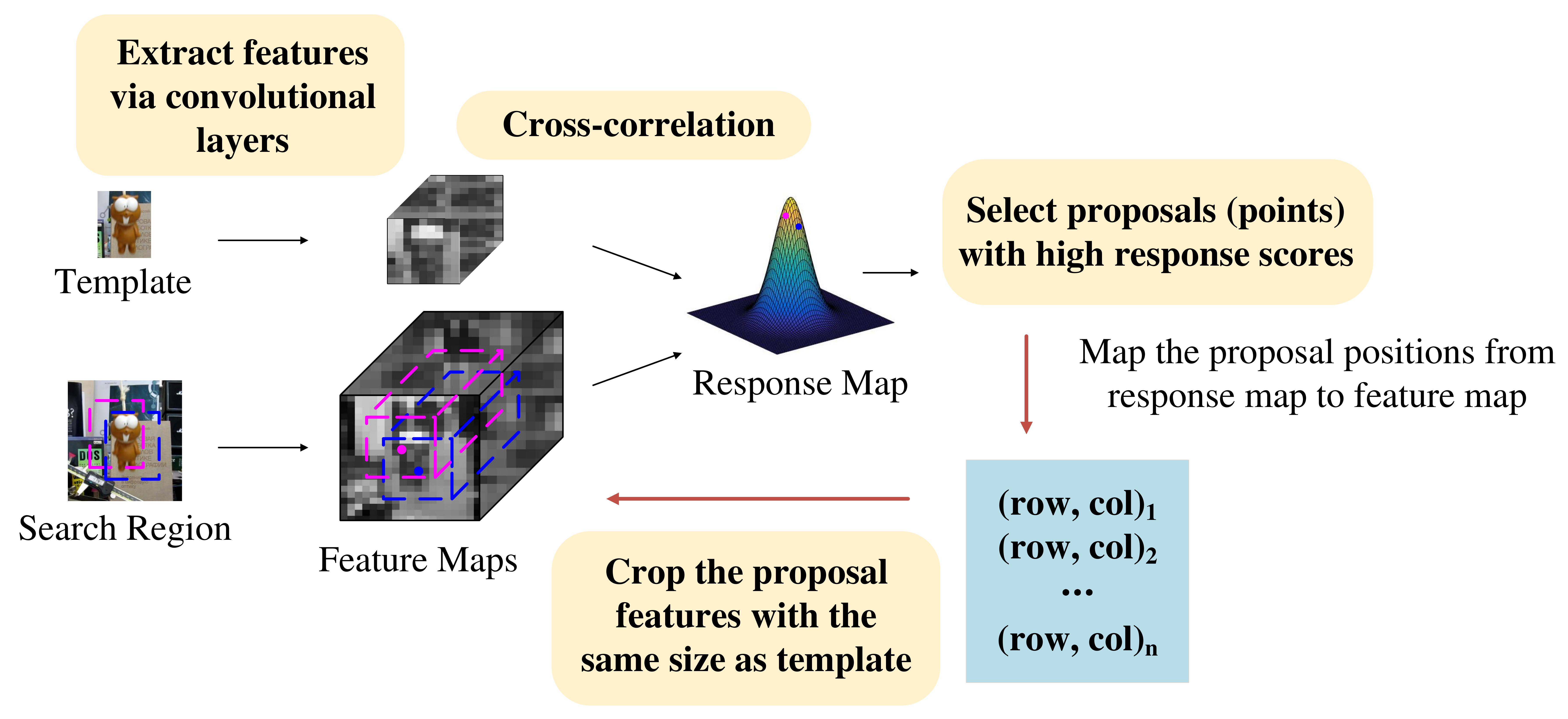}
	\caption{An illustration of the proposed fast proposal selection strategy. In this example, the purple and blue points in the response map denote the similarities for the corresponding proposals in the search region. We crop their features (corresponding to the purple and blue rectangular solids, respectively) from the feature map of the search region. Best viewed in color.}
	\label{fig_fast proposal selection}
\end{figure}

As mentioned above, we aim to filter out the irrelevant and trivial proposals far away from the object center, which can effectively reduce the redundant computation for proposal feature extraction. Although the matching-based tracking method (such as SiamFC \cite{siamesefc}) is sensitive to the changes in object appearance and contexts, it can be effectively used as a coarse pre-estimator. Such a pre-estimator can  identify irrelevant and trivial proposals by comparing them with the initial object appearance. Hence, taking advantage of the high computational efficiency of the fully-convolutional Siamese network, we select the proposals that have high confidence scores to make further evaluation via the subsequent LSTM network.

It is worth pointing out that, different from our previous work \cite{oa-lstm}, we directly crop the features of the selected proposals from the feature map of the search region at the last convolutional layer. As depicted in \cref{fig_fast proposal selection}, a score value in the final response map corresponds to a sub-window in the search region. Thus, we can crop the feature of a proposal by locating its corresponding position in the search region, where the size of features is the same as that of the template features. Then, we feed high-quality proposals (\emph{i.e.}, the selected proposals with high confidence scores) to the online trained LSTM network to perform fine estimation.

This fast proposal selection strategy avoids a mass of redundant computation for the trivial proposals and enables the feature extraction for all the proposals to be performed in a single convolutional forward pass. Such a manner efficiently accelerates the conventional classification-based tracking framework. Note that this proposal selection strategy is adopted to optimize the computational efficiency of  proposal feature extraction, while the following LSTM network is proposed to finely detect the object from the selected proposals with the high adaptability to constantly changing target appearance and contexts. Both components are tightly coupled to promote the tracking performance in both speed and accuracy, especially in challenging scenes.

\subsection{Object-Adaptive LSTM Network} \label{lstm}

\subsubsection{LSTM Network for classification}

Different from the existing classification-based tracking methods \cite{mdnet,sanet}, which simply train the fully-connected layers as a classifier, in this paper we apply an online LSTM  network to visual tracking for classification. As an alternative RNN, the LSTM network inherits the powerful capability of RNNs in modeling sequential data by memorizing the previous input information. In particular, the introduction of the forget mechanism enables the LSTM network to not only capture long-range temporal dependencies, but also ignore distracting information. Hence, the proposed LSTM classification network, which is designed to suit  the visual tracking task, can adapt to the temporally changing object appearance and discriminate the tracked target from the distractors (such as similar objects in background).

As discussed in \cref{fast proposal}, we can obtain high-quality proposals through the proposed fast proposal selection strategy. Then, these selected proposals are further estimated by the LSTM network using the learned temporal dependencies and memorized historical information. Note that, different from common LSTM networks \cite{re3,Recurrent_filter,memtrack} that take a sequence as an input and combine the hidden states of several timesteps as an output,  our LSTM network takes a batch of proposal features in the current frame and the previously estimated LSTM state as inputs, and then estimate a classification result for each proposal features in each frame. The classification result is solely derived from the calculation of the current timestep. After finishing the estimation for the current frame, we choose the LSTM state corresponding to the estimated target state as a new reliable  object representation model, which stores temporal target information and is used in next estimation.

\begin{figure*}[!tb]
	\centering
    \includegraphics[width=1.0\textwidth]{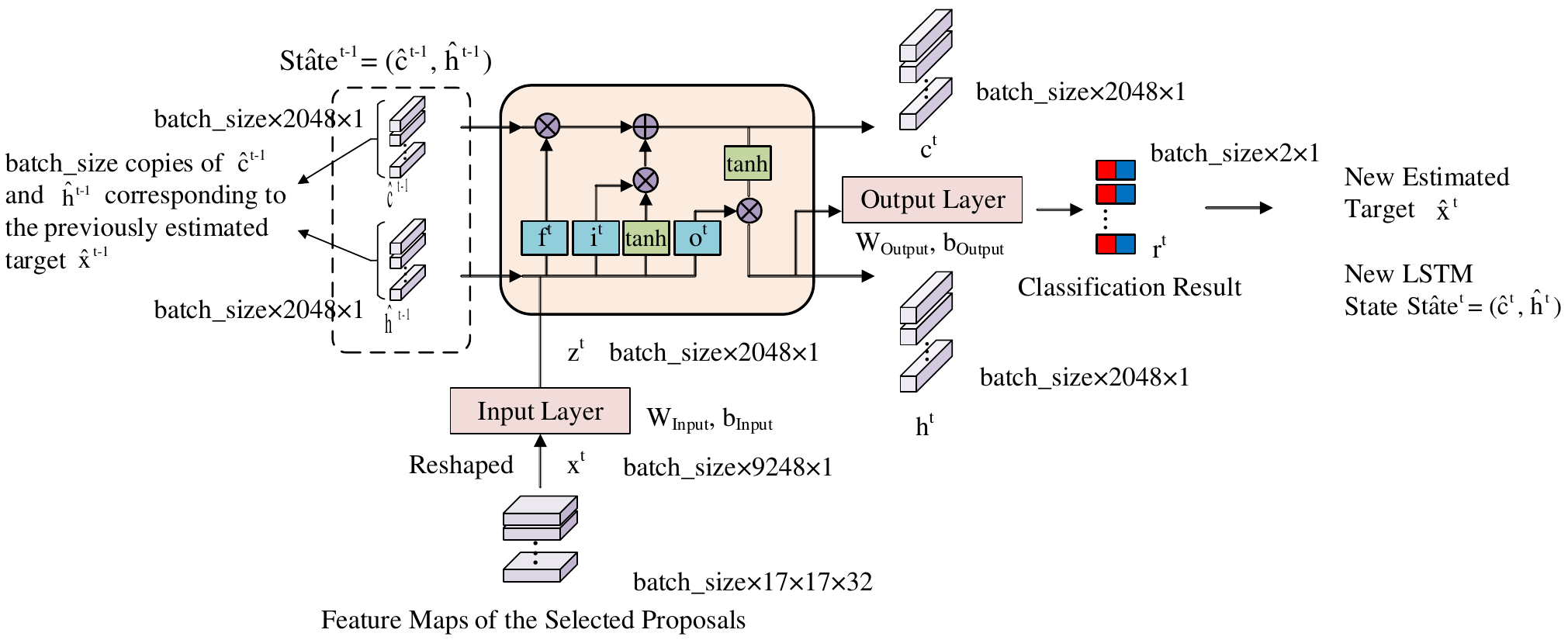}
	\caption{The architecture of the proposed LSTM network. $\hat{c}^{t-1}$ and $\hat{h}^{t-1}$ are the cell and hidden states of the previously estimated target, which together compose the the previously estimated LSTM state $\hat{State}^{t-1}$. $x^t$ is the reshaped feature vector of a $17 \times 17 \times 32$ proposal feature map. $z^t$ is the transfromed feature vector of $x^t$ by the input layer. $c^t$ and $h^t$ are the generated cell and hidden states corresponding to $x^t$. $r^t$ is the classification result. $f^t$, $i^t$ and $o^t$ denote the parameters of forget gates, input gates and output gates in the LSTM blocks, respectively. $W_{Input}$, $b_{Input}$, $W_{Output}$ and $b_{Output}$ respectively represent weight matrices and bias vectors of the input and output layer. In practice, the new estimated LSTM state $\hat{State}^t=(\hat{c}^{t}, \hat{h}^{t})$ corresponding to the new estimated target  $\hat{x}^t$ is fed to the next time step, which allows the information of object representation to propagate through time.}
	\label{fig_lstm}
\end{figure*}

\subsubsection{Forward Pass}
As depicted in \cref{fig_lstm}, the internal architecture of our LSTM blocks is a standard model, while the input layer and the output layer are modified to classify the feature maps of selected proposals. To obtain suitable inputs for our LSTM blocks (vectors in $\mathbb{R}^{n}$, where $n$ is the number of LSTM units), each feature map of selected proposals is directly reshaped to a vector $x^t \in \mathbb{R}^m$. The subsequent input layer is implemented using a fully-connected layer with a weight matrix $W_{Input} \in \mathbb{R}^{ m\times n}$ and a bias vector $b_{Input} \in \mathbb{R}^{n}$, which transforms $x^t \in \mathbb{R}^m$ to $z^t \in \mathbb{R}^{n}$.  The inputs of LSTM blocks in the $t^{th}$ frame consist of three components, \emph{i.e.},  the transformed proposal feature vector $z^t$, the estimated cell $\hat{c}^{t-1}$ and hidden states  $\hat{h}^{t-1}$ in the ${(t-1)}^{th}$ frame. Both $\hat{h}^{t-1}$ and $\hat{c}^{t-1}$ store the previous target information. For brevity, we denote the internal LSTM state in the $t^{th}$ frame by a tuple $State^t=(c^{t}, h^{t})$. Hence, the LSTM blocks take the feature vector $z^t$ and the previously estimated LSTM state $\hat{State}^{t-1}$ as inputs. Note that in the first frame, given the annotation, we can obtain the initial LSTM state ${State}^{1}$ by passing the initial target feature $x^1$ through the LSTM network. Thus, we can start the online tracking process from the second frame using   ${State}^{1}$.

The parameters of input gates $i^t$ and output gates $o^t$ in LSTM blocks control the writing and reading for new target information. The parameters of forget gate $f^t$ control to ignore the useless information such as the background or distractors. The LSTM blocks calculate corresponding cell $c^{t}$ and hidden states $h^{t}$ for each feature vector $z^t$, according to the previously estimated LSTM state $\hat{State}^{t-1}$. Note that our goal is to classify each proposal features, so we use a fully-connected layer with a weight matrix $W_{Output} \in \mathbb{R}^{ n\times 2}$ and a bias vector $b_{Output} \in \mathbb{R}^{2}$ and a following softmax operation to implement the output layer.

By comparing the historical target information  stored in $\hat{State}^{t-1}$ with each proposal feature vector $x^t$, our LSTM network can generate a corresponding new LSTM state $State^t$ (\emph{i.e.}, $State^t=(c^{t}, h^{t})$, which stores the representation information of $x^t$) and the classification result $r^t \in \mathbb{R}^{2}$ (\emph{i.e.}, $r^t=(p^+(x^t), \  p^-(x^t))^T$,  where  $p^+(x^t)$ and  $p^-(x^t)$ are the positive and negative scores of $x^t$). The tracking result is determined by choosing the proposal with the maximum $p^+(\hat{x}^t)$. Its corresponding LSTM state $\hat{State}^{t}$ is considered to represent the optimal target state and used for the next estimation. In online tracking, $\hat{State}^{t}$ maintains an internal object representation model, which can be dynamically updated while receiving new object features. The proposed LSTM network learns to classify the input proposal features $x^t$ according to the previously estimated LSTM state  $\hat{State}^{t-1}$. Specifically, the forward pass of the proposed LSTM network can be calculated with \cref{f1,eq_inputGate,eq_forgetGate,eq_outputGate,f2,f3,f4}.
{
\begin{align}
&\text{Input Layer: } &z^t&= W_{Input}^T x^t+b_{Input}
\label{f1}
\\
&\text{Input Gate: } &i^t &= \sigma(U_\iota z^t + V_\iota \hat{h}^{t-1} +b_\iota)
\label{eq_inputGate}
\\
&\text{Forget Gate: } &f^t &= \sigma(U_\nu z^t + V_\nu \hat{h}^{t-1} +b_\nu)
\label{eq_forgetGate}
\\
&\text{Output Gate: } &o^t &= \sigma(U_\omega z^t + V_\omega \hat{h}^{t-1} +b_\omega)
\label{eq_outputGate}
\\
&\text{Cell: } &c^t &= f^t\odot \hat{c}^{t-1}+i^t \odot \tanh(U_c z^t + V_c \hat{h}^{t-1} +b_c)
\label{f2}
\\
&\text{Cell Output: } &h^t&=o^t \odot \tanh(c^t)
\label{f3}
\\
&\text{Output Layer: } &r^t&=\text{Softmax}(W_{Output}^T h^t+b_{Output})
\label{f4}
\end{align}
}

\noindent {where $i^t$, $f^t$ and $o^t$ denote the parameters of input gates, forget gates  and output gates in the LSTM blocks, respectively.  $U$, $V$ are the  weight matrices and $b$ is the bias vector. The subscript $\iota$, $\nu$, $\omega$ and $c$ respectively refer to the  input gates, forget gates, output gates and LSTM cells. `$\odot$' represents the element-wise product. $\tanh$ and $\sigma$  respectively denote the  hyperbolic tangent activation function and sigmoid activation function. $\text{Softmax}(\cdot)$ represents the softmax activation function.}

\subsubsection{Backward Pass}
We aim to sufficiently utilize the sequence-specific information to track an arbitrary object and avoid the risk of over-fitting to the datasets from the visual tracking domain. Thus, we adopt an online learning strategy to train the LSTM network for the visual tracking task. Particularly, during the training process in the $t^{th}$ frame, instead of feeding a sequence of training data to the LSTM network as done in \cite{re3,Recurrent_filter,memtrack}, we use the previously estimated LSTM state $\hat{State}^{t-1}$ and the training samples $S^t$ drawn from the current frame to train a per-object classifier. In this manner, the LSTM network learns to distinguish the object from the background in accordance with the previously memorized object information. The training loss is directly derived from the classification results. Thus, it does not need to propagate through noisy intermediate timesteps, which can  accelerate the convergence of the LSTM network.

Specifically, in the $1^{st}$ frame, we pass the initial target feature $x^1$ through the LSTM network and obtain the initial LSTM state ${State}^{1}=({c}^{1}, {h}^{1})$. Then, we use ${State}^{1}$ and training samples $S^1$  generated around the original target position to train the LSTM network. In the $t^{th}$ frame, we generate the training samples $S^t$ according to the estimated target state. The LSTM network is updated using $S^t$ and the  previously estimated LSTM state $\hat{State}^{t-1}$ to obtain online adaptability to the temporally changing object appearance and contexts. We use the cross-entropy loss function $\mathcal{L}$ for training. The backward pass in the training process can be calculated with \cref{b1,b2,b3}.
\begin{align}
\epsilon^t_r &\overset{def}{=}\frac{\partial\mathcal{L} }{\partial r^t} \  \frac{\partial  r^t }{\partial \text{Softmax}(\cdot)}
\label{b1}
\\
\epsilon^t_h &= W_{Output} \epsilon^t_r
\label{b2}
\\
\epsilon^t_c &=  {(o^t)}'\tanh(c^t)\epsilon^t_h +o^t\tanh'(c^t)\epsilon^t_h
\label{b3}
\end{align}

\noindent  where $\epsilon^t_r$ is defined as  the derivative of loss function $\mathcal{L}$ with respect to the softmax activation function $\text{Softmax}(\cdot)$, \emph{i.e.}, the derivative of the softmax cross-entropy loss function. $\epsilon^t_h$ and $\epsilon^t_c$ denote the derivatives of loss function $\mathcal{L}$  with respect to $h^t$ and $c^t$, respectively. ${(o^t)}'$ refers to the derivative of $o^t$ with respect to $c^t$, \emph{i.e.}, ${(o^t)}'=\frac{\partial o^t }{\partial c^t}$.  $\tanh'(\cdot)$ represents  the derivative of the hyperbolic tangent activation function.

\subsection{Data Augmentation with GAN}

To learn a robust classifier that can effectively discriminate the object from the background in challenging scenes, the online training of the LSTM network requires adequate labelled training data. However, since only one object is provided despite the comparatively broad background for the visual tracking task, the number of positive samples is relatively small and is far less than the number of negative samples. The problems of sample inadequacy and positive-negative class imbalance will hinder the online training of the LSTM network and need to be tackled properly. Compared with our previous work \cite{oa-lstm}, we present a data augmentation strategy based on GAN \cite{gan2014} to generate diverse positive samples in the image space. The proposed strategy enriches the available training data and thus effectively boosts the performance of the proposed method.

In this paper, we adopt a recently developed generative adversarial model \cite{dcgan} (DCGAN) for the training stability. Since the tracking method needs to track an arbitrary object, it is difficult to pre-train a general sample augmenter.
Therefore, during online tracking, we train GAN in the first frame to learn the original target appearance and then update it with real sampled images in the subsequent frames to  effectively capture temporarily changing target appearance.

In the generative adversarial learning process, a real image $x$ of positive sample drawn from the frames obeys the distribution $P_{img}(x)$. The model contains a generator $G$ to learn this real data distribution and a discriminator $D$ to distinguish the real images from the generated images. The generator takes a noise variable $P_{noise}(z)$ as the input and it outputs an image $G(z)$ that approximates the real image $P_{img}(x)$. The discriminator $D$ takes both $P_{img}(x)$ and $G(z)$ as inputs and outputs their classification probability. On one hand, we train $D$ to maximize the classification probability of assigning the correct labels to both the real images and generated images. On the other hand, we train $G$ to maximize the probability of $D$ making a mistake, \emph{i.e.}, to minimize the classification probability of $G(z)$ assigned with the correct label. Hence, $D$ and $G$ play a two-player minimax game with the following function:
\begin{align}
	\label{gan}
	\min \limits_G \max \limits_D F(D,G) = &\mathbb{E}_{x \sim P_{img}(x)}[\log D(x)]  \\
	&+\mathbb{E}_{z \sim P_{noise}(z)}[\log(1-D(G(z)))]. \nonumber
\end{align}

By the adversarial training, $D$ and $G$ boost their respective performance from each other until $D$  cannot distinguish the differences between the real images and the generated ones. In this way, $G$ effectively learns the real data distribution $P_{img}$. The generated images closely approximate the real images.

\begin{figure}[!tb]
	\centering
	\includegraphics[width=0.9\textwidth]{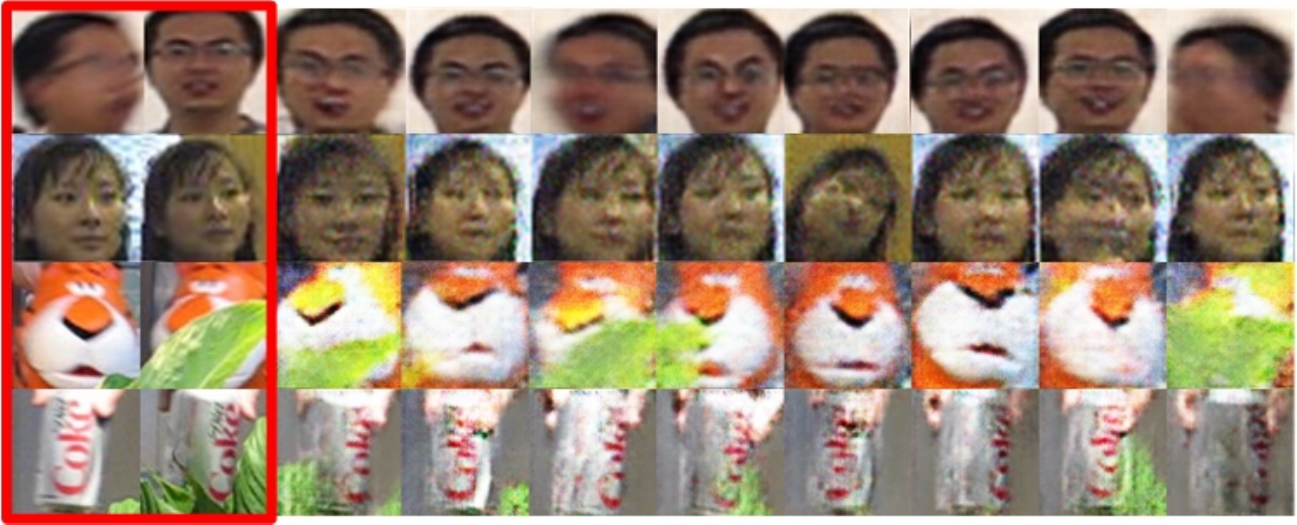}
	\caption{The left two columns in the red rectangle are real images of positive samples. The right eight columns are the generated positive samples with GAN on the four sequences from the OTB dataset (from top to down: \emph{Boy}, \emph{Girl}, \emph{Tiger1} and \emph{Coke}, respectively). Best viewed in color.}
	\label{fig_generated samples}
\end{figure}

\cref{fig_generated samples} presents the real images of positive samples and the generated  positive samples based on GAN. We take real images of positive samples as $P_{img}(x)$, which are drawn around the estimated target position from video frames. The noise variable $P_{noise}(z)$ is randomly generated. After the adversarial learning process, we apply the learned generator $G$ to sample a number of positive samples $G(z)$. Then, we augment the training data of the LSTM network with these generated positive samples. By this way, the problem of class imbalance is alleviated. As shown in \cref{section_internal comparison}, this data augmentation strategy facilitates the online training of the LSTM network and improves the tracking accuracy of the proposed method.

\subsection{Discussions}
It is worthy mentioning that the proposed  method exploits but differs from the previous works, including SiamFC \cite{siamesefc}  and DCGAN \cite{dcgan}.

In this paper, we propose a novel and fast proposal selection strategy
to accelerate the LSTM classification network. Specifically, we take advantage of
the response map of the matching-based tracking method (SiamFC is used in this paper) to select high-quality proposals and directly obtain
the proposal features from the feature map of search region. Such a strategy
effectively avoids the heavy computation for proposal feature extraction in the
classification based tracking framework. In contrast,  SiamFC adopts an offline pretrained model, which directly outputs the proposal with the highest response score as the tracked result.
 In other words, SiamFC does not perform object-adaptive proposal re-estimation and  inherently lacks online adaptability.

The proposed data augmentation technique is based on DCGAN. However, DCGAN \cite{dcgan} is trained on various image datasets for general image representations, while our data augmenter is learned online with sequence-specific information, which better suits for the visual tracking task. In addition, we incorporate it into our recurrent tracking model to facilitate the training of the proposed object-adaptive LSTM network.

\section{Online Tracking Algorithm}
\label{section_algorithm}
\subsection{Online Training of the Network Model}
As discussed in \cref{fast proposal},
the Siamese network (\emph{i.e.}, SiamFC \cite{siamesefc}) used in our fast proposal selection is trained offline using pairs of images taken from the ILSVRC15 \cite{imagenet} dataset, which avoids the risk of over-fitting to the datasets in the visual tracking domain.
Since the Siamese network is used as a coarse pre-estimator, we directly apply the pre-trained Siamese network to select the high-quality proposals  without online updating. In the following, we introduce the online training of the LSTM network, which is designed to further estimate the selected proposals by exploiting temporal dependencies.

Given the annotated first frame, we feed the LSTM network with the original target appearance to initialize the LSTM state. Then, we draw the positive and negative samples around the original target position with the normal distribution. We use the training samples from the first frame and the original LSTM state to train the LSTM network as stated in \cref{lstm}. In the subsequent frames,
we update the LSTM network using the training samples drawn around the estimated target position and the previously estimated LSTM state. Through online learning, the LSTM network is encouraged to discriminate the object from the background according to the previously estimated LSTM state which stores the historical information of object representation. Besides, due to its intrinsic recurrent structure, the LSTM network can dynamically update its recurrent parameters during the forward passes. Thus, the model of object representation stored in the LSTM state is constantly updated as new inputs of proposal features are received.

\subsection{Online Tracking Using OA-LSTM-ADA}

\begin{algorithm}[tb!]
	\renewcommand\baselinestretch{1}\selectfont
	\caption{Tracking algorithm of OA-LSTM-ADA}
	\begin{algorithmic}[1]
		\label{algorithm1}
		\vspace*{0.04in}
		\REQUIRE
		Original target state $x^1$, similarity learning function  $\mathcal{F}$, predefined threshold $\theta$
		\ENSURE
		Estimated target state $\hat{x}^t$
		\STATE Initialize the Object-Adaptive LSTM network using $x^1$;
		\STATE Sample training data $s^1_+$ and $s^1_-$ from the 1\textsuperscript{st} frame,\\
		$S^1 \leftarrow \{s^1_+\} \cup \{s^1_-\}$;
		\STATE Train the Object-Adaptive LSTM network using $S^1$;
		\STATE Train GAN with the positive samples $s^1_+$;
		\REPEAT
		\STATE Apply the similarity learning function $\mathcal{F}$ to obtain a confidence map $\mathcal{M}$;
		\STATE Select $N$ high-score proposals $\{ {x}_i^t \}^N_{i=1}$ from $\mathcal{M}$;
		\STATE Evaluate $\{  {x}_i^t \}^N_{i=1}$ with the previously estimated LSTM state $ {\hat{State}}^{t-1}$ to obtain their positive scores $\{p^+( {x}_i^t )\}^N_{i=1}$;
		\STATE Find the tracked result by $ {\hat{x}}^t=\mathop{\arg\max}_{ {x}_i^t} p^+(  {x}_i^t )$;
		\STATE Set the optimal LSTM state $ {\hat{State}}^t$ corresponding to $ {\hat{x}}^t$;
		\IF{ $p^+(\hat{x}^t)> \theta $ }
		\STATE Sample training data $s^t_+$ and $s^t_-$ by using the hard negative mining technique, $S^t \leftarrow \{s^t_+\} \cup \{s^t_-\}$;
		\STATE Take $\{s^1_+, ..., s^t_+\}$ as the inputs, and generate diverse positive samples $g^t_+$ using GAN, $S^t \leftarrow S^t \cup \{g^t_+\}$;
		\STATE Update the LSTM network using $S^t$;
		\ENDIF
		\UNTIL{end of sequence}
	\end{algorithmic}
\end{algorithm}

Our online tracking algorithm of the Object-Adaptive LSTM network with Adversarial Data Augmentation (OA-LSTM-ADA) is presented in \cref{algorithm1}. The similarity learning function $\mathcal{F}$ refers to the Siamese network \cite{siamesefc} used in the fast proposal selection step (see \cref{fast proposal}). $\mathcal{F}$ can be regarded as a general function that calculates the similarities between the target template and the candidate patches. $\theta$ is a predefined threshold for the online update of the LSTM network. When the positive score of the estimated target state exceeds $\theta$, the tracked result is considered to be reliable and it can be used for the sampling of training data.

In the first frame, we initialize the LSTM network using the original target state $ {x}^1$ and train the network with the training data $S^1$ drawn from the first frame. The drawn positive data $s^1_+$ are taken as the input real images for the initial training of GAN. After the initial training, the generator of GAN coarsely learns the appearance representation of the object.

In the subsequent $t^{th}$ frame, we firstly pre-evaluate the densely sampled proposals  with the similarity learning function $\mathcal{F}$ and select high-quality ones to feed to the following LSTM network. Then, the selected proposals are estimated by the LSTM network according to the previously estimated LSTM state $ {\hat{State}}^{t-1}$. We obtain the positive scores and negative scores of the selected proposals and treat the one with the maximum positive score to be the tracked result $ {\hat{x}}^t$. The optimal LSTM state $ {\hat{State}}^{t}$ corresponding to $ {\hat{x}}^t$ is accordingly updated and will be used for the estimation of target state in the next frame.

When the positive score of the estimated target state exceeds $\theta$, we perform the update procedure. In order to improve the robustness of the LSTM network to deal with the similar objects in the background, we apply the hard negative mining technique \cite{hard_negative_mining} to draw training samples $S^t$. Note that we can directly use the confidence map $\mathcal{M}$ to select hard negative samples and do not require the extra computational cost for sample evaluation. This technique makes the LSTM network more discriminative when the background contains similar objects to the tracked target.

Taking the positive samples $\{s^1_+, ..., s^t_+\}$ as the input real images, we use GAN to generate diverse positive samples $g^t_+$ and augment the training data $S^t$. Therefore, the LSTM network is updated with the augmented training data $S^t$ that contain adequate positive samples and hard negative samples. This strategy  provides the LSTM network with high adaptability to the temporarily changing object and background.

\section{Experiments}
\label{section_experiment}
To evaluate the performance of the proposed tracking method, we conduct extensive experiments on four public tracking benchmarks, \emph{i.e.}, OTB (including OTB-2013 \cite{otb13} and OTB-2015 \cite{otb15}), TC-128 \cite{tc128}, UAV-123 \cite{uav123} and VOT-2017 \cite{vot2017}. In \cref{section_ex_implementation}, we present the implementation details and parameter settings used in our experiments. In \cref{section_otb}, we evaluate our tracker on the OTB dataset by providing internal comparison, quantitative comparison, attributed-based comparison and qualitative comparison. In \cref{section_tc128}, \cref{section_uav123} and \cref{section_vot}, we conduct the evaluation on the TC-128, UAV-123 and VOT-2017 datasets respectively, showing the results of quantitative comparison with several state-of-the-art trackers.

\subsection{Implementation Details and Parameter Settings}
\label{section_ex_implementation}
Our tracker, OA-LSTM-ADA, is implemented in Python using TensorFlow \cite{tensorflow}. It runs at an average speed of 32.5 fps  with a 2.7 GHz Intel Core i7 CPU with 16 GB RAM and an NVIDIA GeForce GTX Titan X GPU. In the proposed fast selection strategy, we utilize the matching-based tracking method, \emph{i.e.}, SiamFC-3s \cite{siamesefc} (the version searching over 3 scales instead of 5 scales). The template used in the Siamese network is the original object appearance in the first frame. We set the size of the Siamese response map to 33 $\times$ 33 without upsampling. To obtain the features of the selected proposals, we crop the feature patches with the size of 17 $\times$ 17 (the same size as the template feature patch) from the feature map (with the size of 49 $\times$ 49) of the search region. Since SiamFC-3s scales the exemplar images and search images with an added margin for context, we set the parameter of context to 0.2 to alleviate the effects of the added context in our classification model. We experimentally select 64 high-quality proposals, which is effective and efficient for a trade-off between performance and speed.

In the proposed LSTM network, we adopt a two-layer LSTM network, each layer of which has 2,048 units. We use the ADAM gradient optimizer \cite{adam} with a softmax cross-entropy loss function and a learning rate of $10^{-5}$. In the proposed data augmentation strategy, we utilize a recent state-of-the-art model (DCGAN \cite{dcgan}) and generate 64 positive samples in each update. In \cref{algorithm1}, the positive score of the estimated target state $p^+(\hat{x}^t)$ is normalized and the threshold parameter $\theta$ for online update of the LSTM network is set to 0.6, which is efficient experimentally. In addition, we conduct all the experiments with the same parameter settings to guarantee the reliability of our experimental results.


\subsection{Evaluation on OTB}
\label{section_otb}
\subsubsection{Dataset and Evaluation Metrics}
\label{otb-metrics}
The OTB-2013 \cite{otb13} dataset consists of 50 fully annotated video sequences with eleven challenging attributes, such as scale variation, illumination variation, occlusion, \emph{etc}. The OTB-2015 \cite{otb15} dataset is the extended version of OTB-2013, which contains the entire 100 fully annotated video sequences with substantial variations.

We adopt the straightforward One-Pass Evaluation (OPE) as the performance evaluation method.
For the performance evaluation metrics, we use precision plots and success plots.
Following the protocol in the OTB benchmark, we  use the threshold of 20 pixels and area under curve (AUC) to present and compare the representative precision plots and success plots of trackers, respectively.

\subsubsection{Internal Comparison}
\label{section_internal comparison}

\begin{figure}[!tb]
	\setlength{\abovecaptionskip}{-0.3em}
	\centering
	\subfigure[Internal Comparison on OTB-2013]{
		\label{OTB-2013-pre-ablation}
		\includegraphics[width=0.4\textwidth]{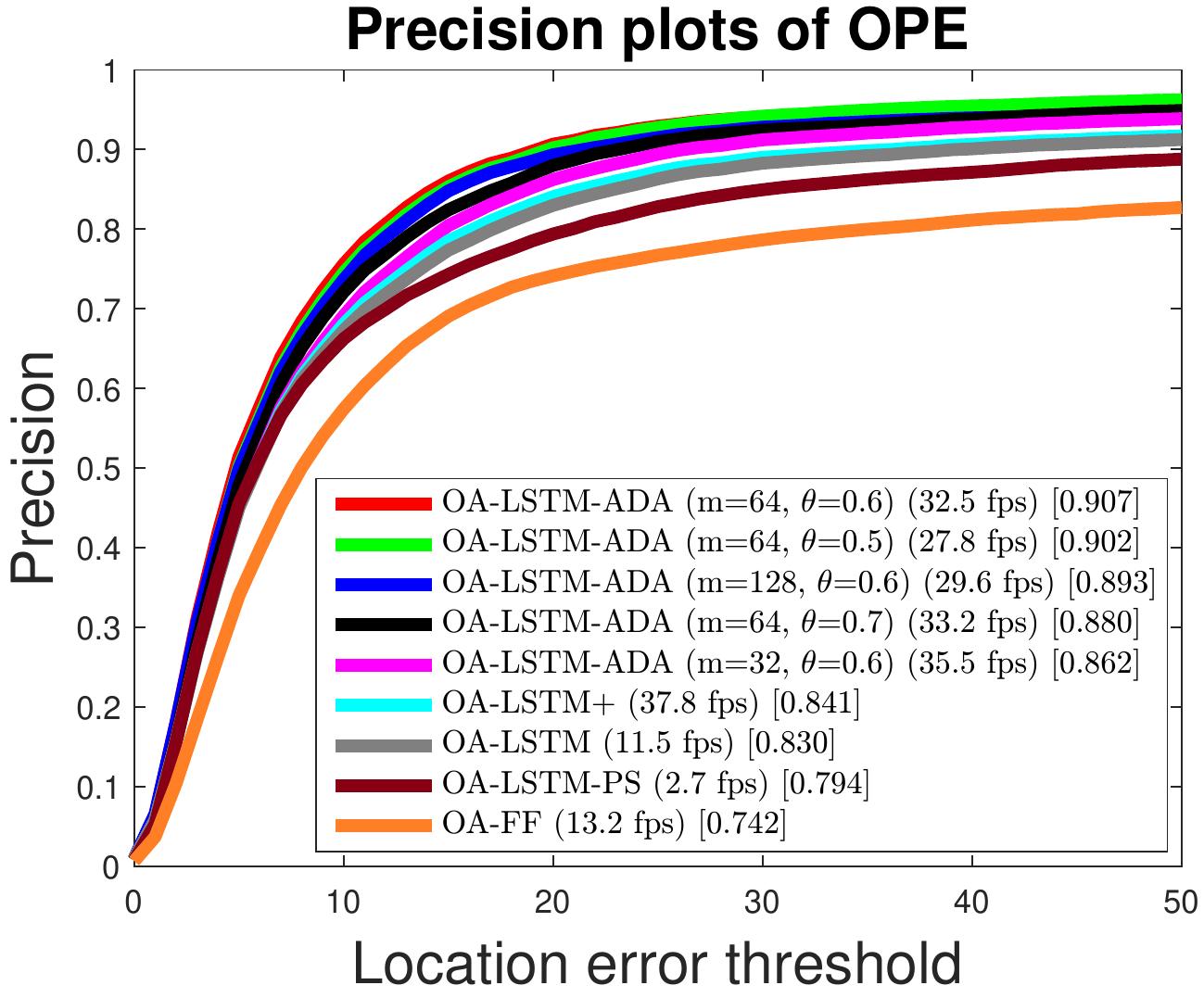}
		\label{OTB-2013-suc-ablation}
		\includegraphics[width=0.4\textwidth]{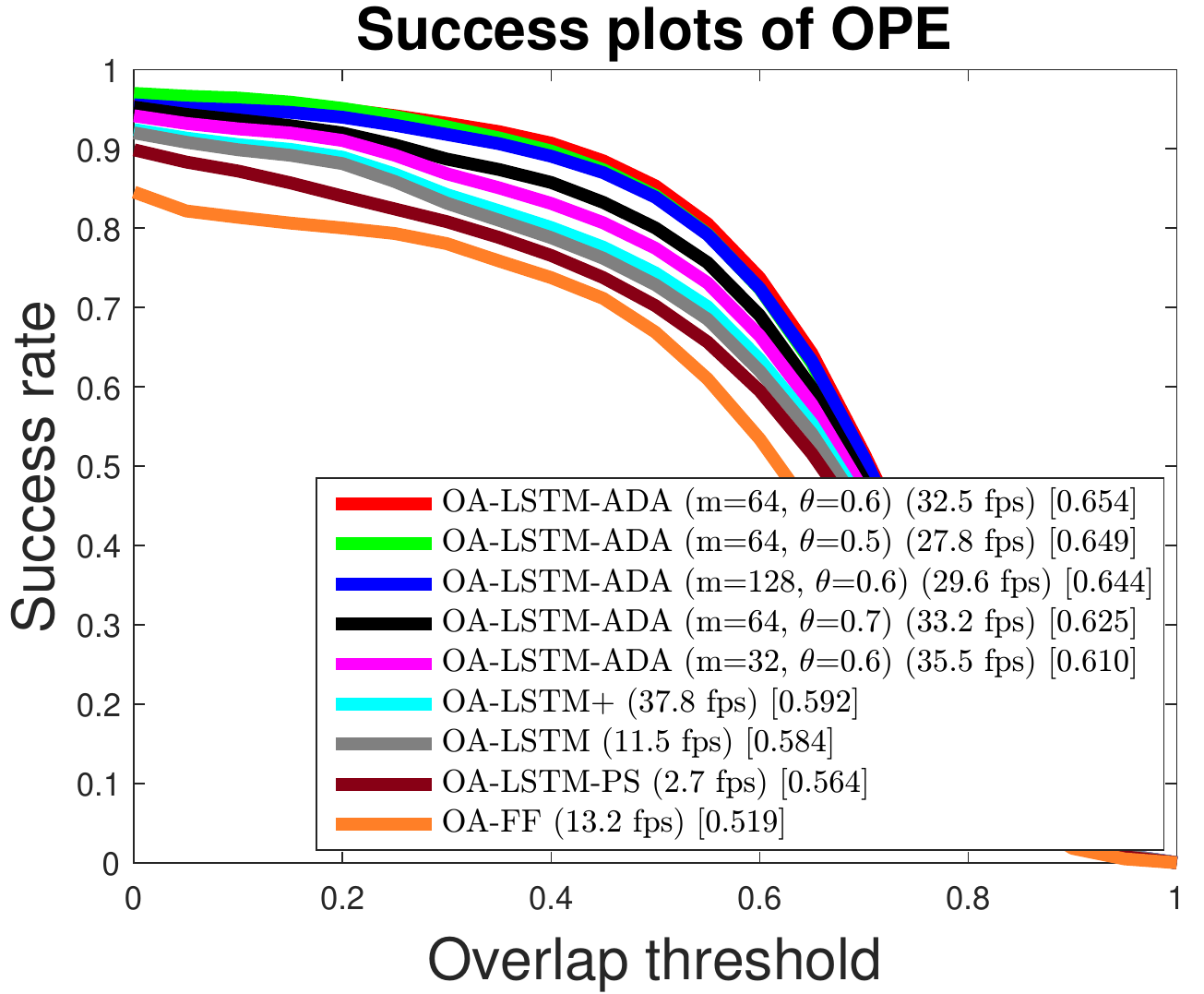}}
	
	\subfigure[Internal Comparison on OTB-2015]{
		\label{OTB-2015-pre-ablation}
		\includegraphics[width=0.4\textwidth]{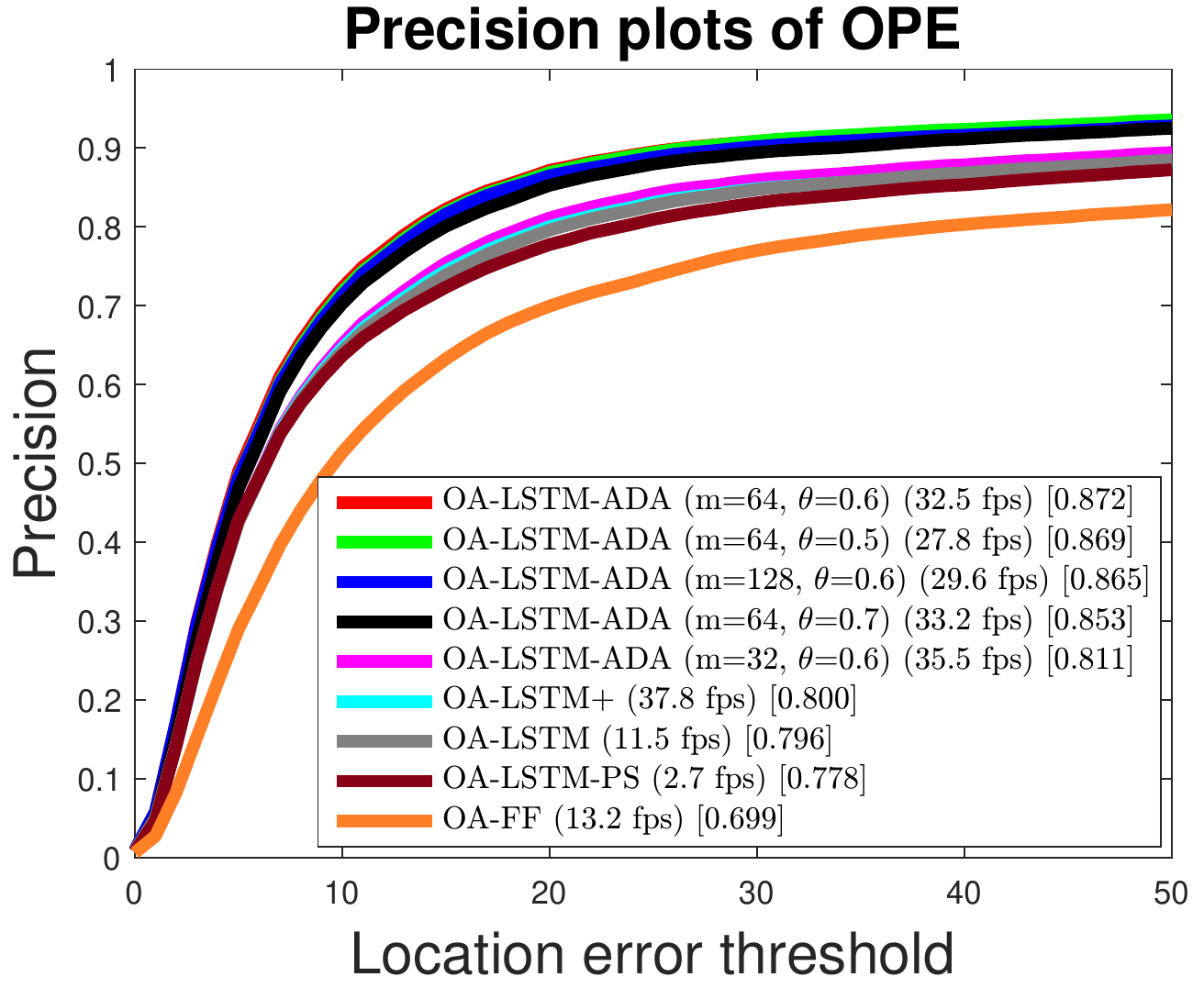}
		\label{OTB-2015-suc-ablation}
		\includegraphics[width=0.4\textwidth]{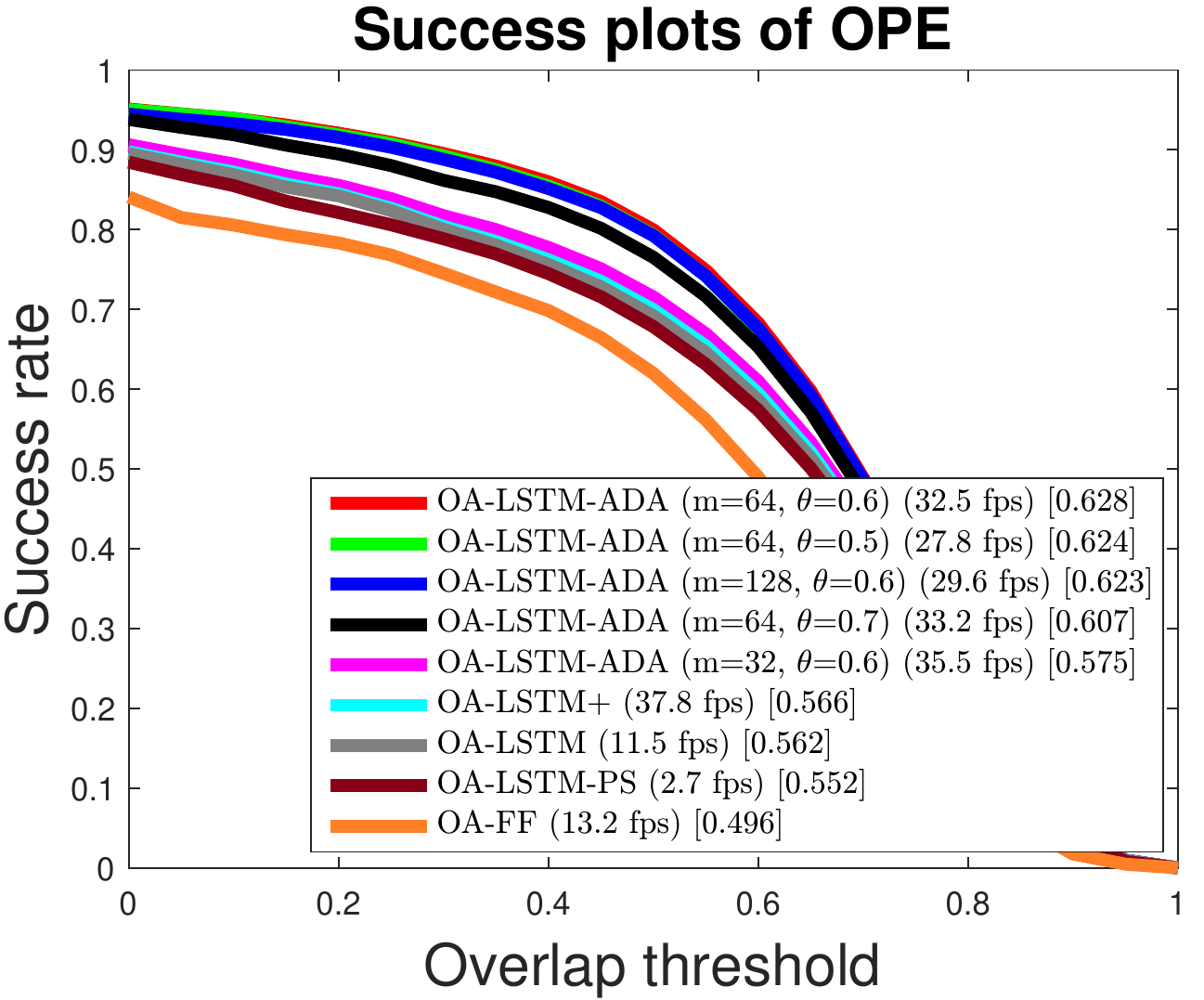}}
	\caption{Results of internal comparison on the (a) OTB-2013 and (b) OTB-2015 datasets. The speeds are presented in the legend.}
	\label{fig_ablation}
\end{figure}

In OA-LSTM-ADA, we adopt a novel object-adaptive LSTM network to utilize time dependencies and memorize the object appearance variations. We also employ the fast proposal selection strategy to improve the computational efficiency. In addition, to facilitate the online training of the LSTM network, we present a data augmentation technique based on GAN. To validate the effectiveness of each component in OA-LSTM-ADA, we investigate its four variants:
\begin{itemize}
	\item OA-FF: a feed-forward variant, where the LSTM network is replaced by the fully-connected layers.
	\item OA-LSTM-PS: a variant without using fast proposal selection, which performs dense sampling and tracks the object via the proposed LSTM network.
	\item OA-LSTM: our previous work \cite{oa-lstm}, which cumbersomely extracts the proposal features by passing the proposal patches through convolutional layers and does not employ the data augmentation technique.
\item OA-LSTM+: an accelerated version of OA-LSTM \cite{oa-lstm}, which directly crops the proposal features from the feature map of search region and does not adopt the data augmentation technique.
\end{itemize}

We evaluate four variants on the OTB-2013 and OTB-2015 datasets and compare their tracking performance with the proposed OA-LSTM-ADA.

As shown in \cref{fig_ablation}, all the variants perform worse than OA-LSTM-ADA in terms of tracking accuracy.
OA-FF simply classifies the selected proposals with the fully-connected layers and it does not effectively capture time dependencies among sequential frames. As a result, OA-FF cannot adapt to the temporarily changing object, and thus it is prone to drift in challenging scenes.
OA-LSTM-PS  is much slower than other methods due to the heavy computational burden caused by dense sampling.
OA-LSTM and OA-LSTM+ show similar tracking accuracy due to the effectiveness of the object-adaptive LSTM network.
However, OA-LSTM+ achieves a higher speed  by directly obtaining the selected proposal features from the big feature map of the search region, which accelerates our original fast proposal selection strategy. This implies that the proposed fast proposal selection strategy effectively reduces the redundant computation for feature extraction and leads to a significant speedup.
OA-LSTM-ADA achieves the best tracking accuracy and satisfactory speed among the compared versions.
This is because that OA-LSTM-ADA employs GAN to augment training data for the online training of the LSTM network, which effectively improves the tracking performance.
Although the speed of OA-LSTM-ADA is slightly lower than that of OA-LSTM+ due to the additional data augmentation technique, OA-LSTM-ADA achieves significant improvements in tracking accuracy by taking advantage of enriched training samples.

Moreover, we further experimentally investigate the influence of the number of selected proposals $m$ and the predefined threshold $\theta$ on the performance and speed of OA-LSTM-ADA. We select a range of values for these two parameters, \emph{i.e.}, $m \in \{32,\  64,\  128\}$ and $\theta \in \{0.5,\  0.6,\  0.7\}$.
The results are given in \cref{fig_ablation}.
As shown in \cref{fig_ablation}, the proposed method with the parameter setting $m=64,\  \theta=0.6$  for OA-LSTM-ADA obtains the best performance among all the parameter settings. While the proposed method with this parameter setting shows slightly slower speed than that with the parameter settings $m=32,\  \theta=0.6$ and $m=64,\  \theta=0.7$, it achieves better trade-off between tracking accuracy and speed. Therefore, we set $m=64,\  \theta=0.6$ for practical efficiency in the following.

\subsubsection{Quantitative Comparison}
\label{quantitative_otb}

\begin{figure}[!tb]
	\setlength{\abovecaptionskip}{-0.3em}
	\centering
	\subfigure[State-of-the-art Comparison on OTB-2013]{
		\label{OTB-2013-pre}
		\includegraphics[width=0.4\textwidth]{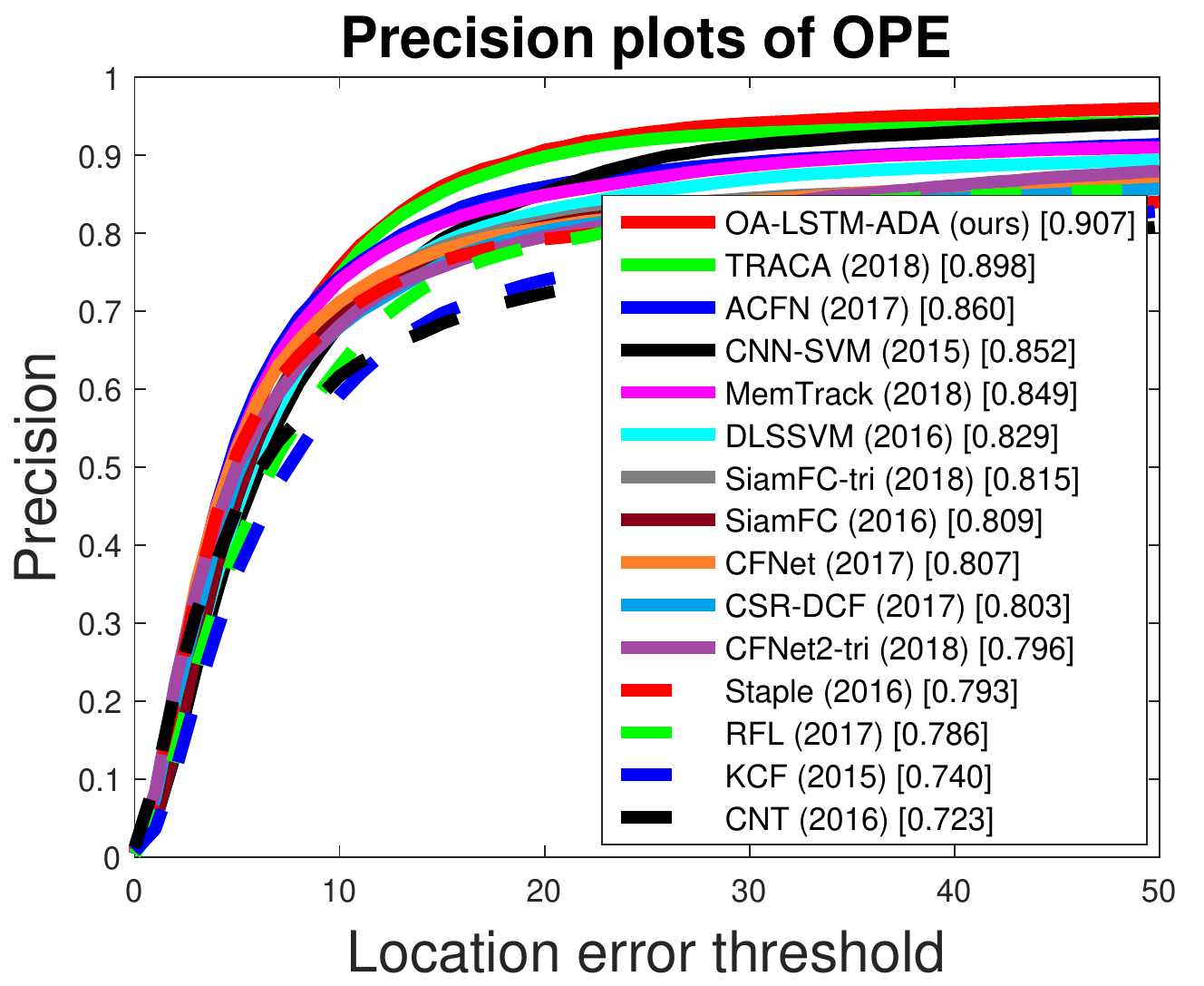}
		\label{OTB-2013-suc}
		\includegraphics[width=0.4\textwidth]{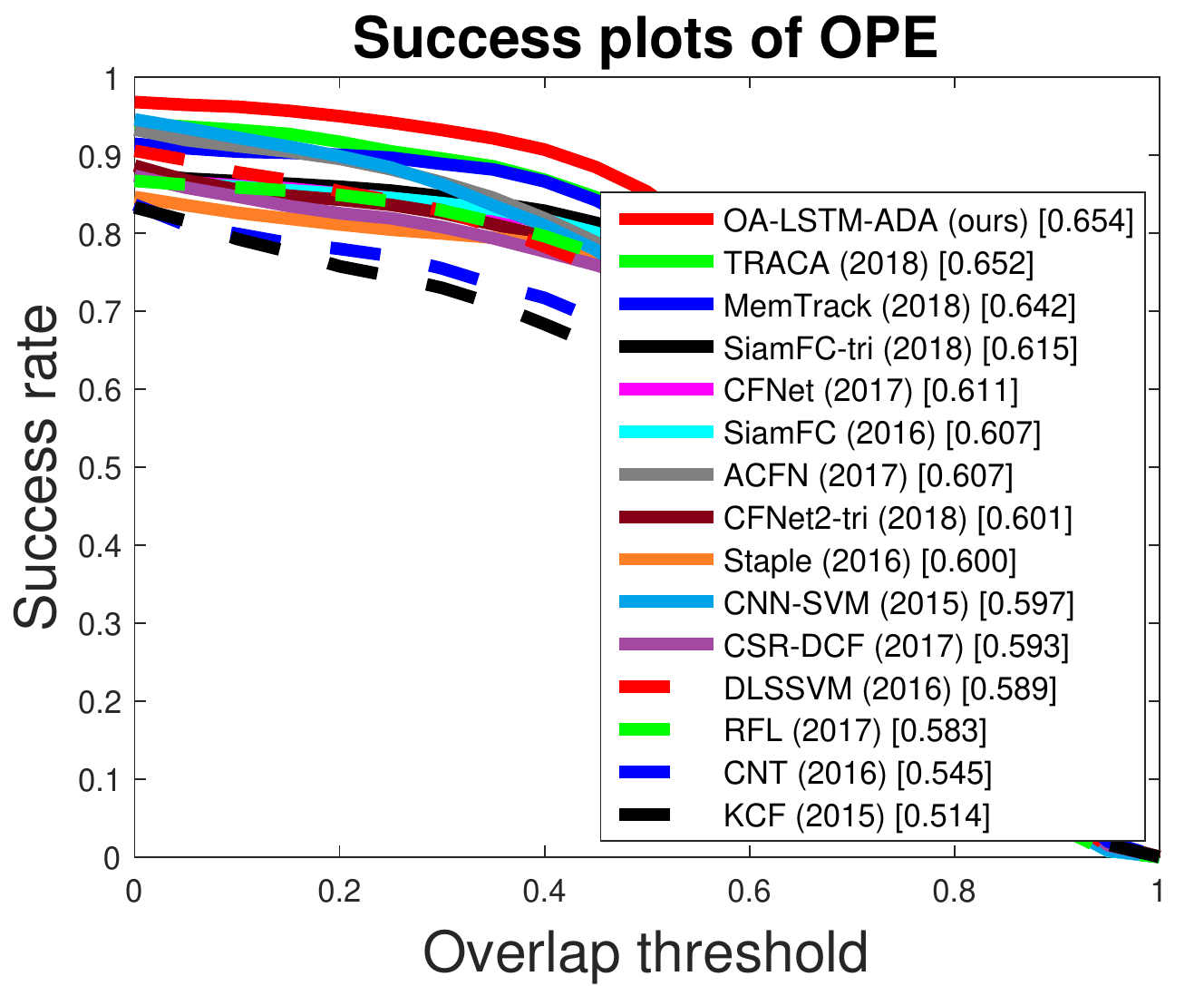}}
	
	\subfigure[State-of-the-art Comparison on OTB-2015]{
		\label{OTB-2015-pre}
		\includegraphics[width=0.4\textwidth]{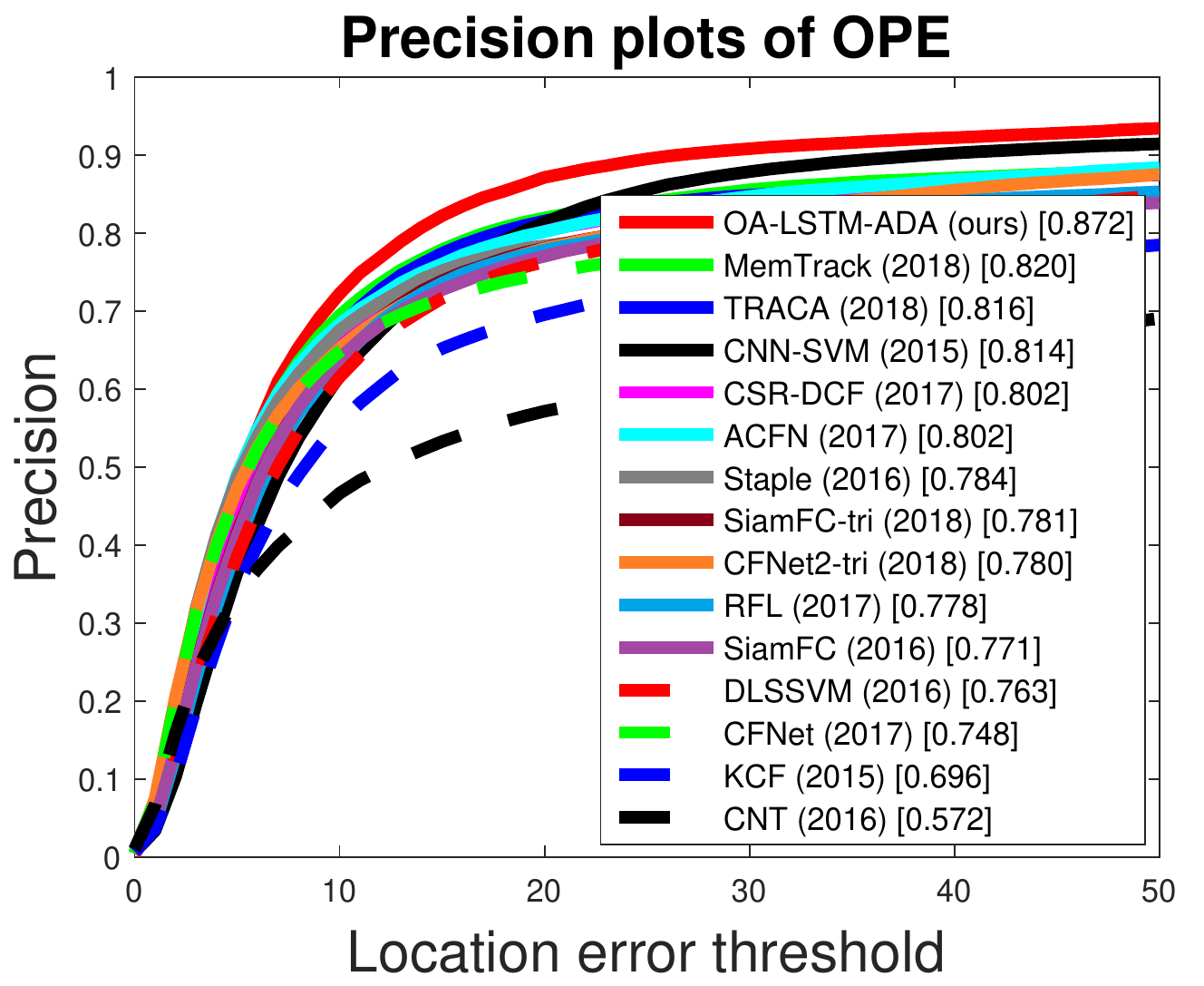}
		\label{OTB-2015-suc}
		\includegraphics[width=0.4\textwidth]{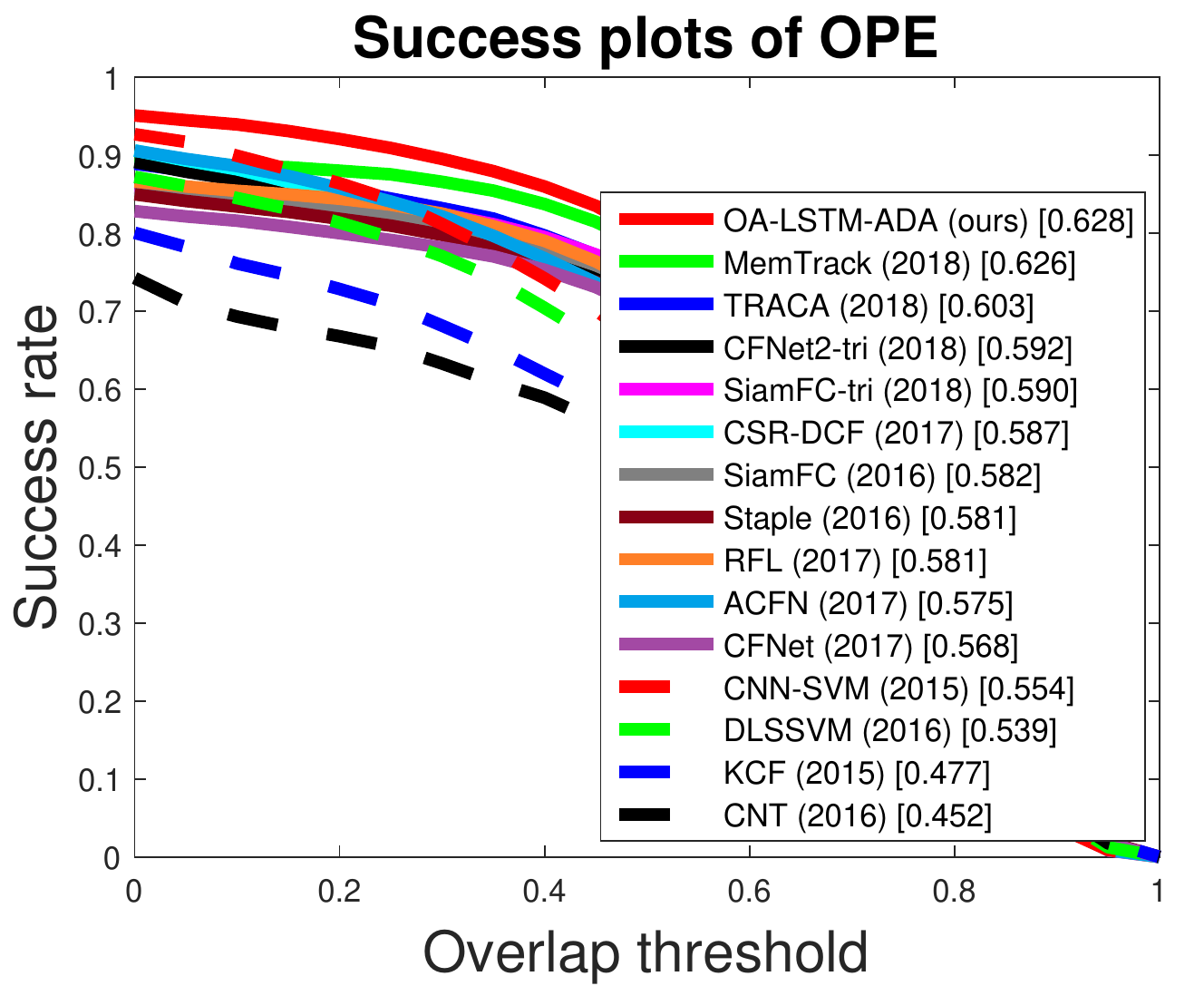}}
	\caption{Precision plots and success plots showing the performance of our OA-LSTM-ADA compared with other state-of-the-art trackers on the (a) OTB-2013 and (b) OTB-2015 datasets.}
	\label{fig_otb}
\end{figure}

As illustrated in \cref{fig_otb}, we compare the precision plots and success plots obtained by our OA-LSTM-ADA and several state-of-the-art trackers including MemTrack \cite{memtrack}, TRACA \cite{TRACA}, SiamFC-tri \cite{Siam-tri}, CFNet2-tri \cite{Siam-tri}, ACFN \cite{acfn}, CNN-SVM \cite{cnn-svm}, DLSSVM \cite{dlssvm}, SiamFC \cite{siamesefc}, CFNet \cite{cfnet}, CSR-DCF \cite{csr-dcf}, Staple \cite{staple}, RFL \cite{Recurrent_filter}, KCF \cite{kcf} and CNT \cite{cnt}. We choose these methods because SiamFC, CFNet, SiamFC-tri  and CFNet2-tri  are Siamese network based tracking methods, which are closely related to our OA-LSTM-ADA (recall that OA-LSTM-ADA utilizes the Siamese network to pre-estimate the densely sampled proposals). MemTrack and RFL also combine the Siamese networks and LSTM networks, but their LSTM networks are used for object template management. Since our tracker adopts deep features for object representation, we choose some representative methods based on deep features, \emph{i.e.}, TRACA, ACFN, CNN-SVM, DLSSVM and CNT. We also choose some state-of-the-art real-time methods based on correlation filters, \emph{i.e.}, CSR-DCF, Staple and KCF.

We can observe that our OA-LSTM-ADA performs favorably among the state-of-the-art trackers on both benchmark versions. Compared with the four Siamese network based trackers, \emph{i.e.}, SiamFC, CFNet, SiamFC-tri  and CFNet2-tri, OA-LSTM-ADA achieves higher tracking accuracy. This fully validates the effectiveness of the proposed novel object-adaptive LSTM network. OA-LSTM-ADA performs  better than MemTrack and RFL with respect to both precision plots and success plots, which demonstrates that our LSTM network is successful in classifying proposals using its memorized target information, compared with the matching-based recurrent trackers. OA-LSTM-ADA also outperforms other deep learning based trackers, \emph{i.e.}, TRACA, ACFN, CNN-SVM, DLSSVM and CNT. This is because that OA-LSTM-ADA not only uses deep features, but also exploits the sequential dependencies in a video and captures the object appearance variations via the LSTM network. Other trackers using hand-crafted features, \emph{i.e.}, CSR-DCF, Staple and KCF, adopt the popular correlation filter tracking framework and achieve state-of-the-art performance. However, these methods achieve worse tracking results than our OA-LSTM-ADA, due to the lack of high-level semantic understanding in challenging scenes. Note that the results of some state-of-the-art methods are directly taken from \cite{Galoogahi} (using the same hardware platform).

\begin{table}[!tb]
	\centering
	\caption{The precision
		score, the AUC (Area Under the Curve) score and
		speed (fps, * indicates the GPU speed, otherwise the CPU speed) on the OTB-2015 dataset. The best and second best results are displayed in red and blue fonts, respectively.} \label{table_otb}
	\setlength{\tabcolsep}{6mm}
	\renewcommand\arraystretch{0.8}
	\scalebox{0.8}{
		\begin{tabular}{cccc}
			\toprule
			Tracker&Precision&AUC&Speed\\
			\midrule
			\textbf{OA-LSTM-ADA}&{\color{red}87.2}&{\color{red}62.8}&32.5*\\
			MemTrack \cite{memtrack}&{\color{blue}82.0}&{\color{blue}62.6}&50.0*\\
			TRACA \cite{TRACA}&81.6&60.3&{\color{blue}101.3*}\\
			CNN-SVM \cite{cnn-svm}&81.4&55.4&1.0*\\
			CSR-DCF \cite{csr-dcf}&80.2&58.7&16.4\\
			ACFN \cite{acfn} &80.2&57.5&15.0*\\
			Staple \cite{staple}&78.4&58.1&50.8\\
			SiamFC-tri \cite{Siam-tri}&78.1&59.0&86.3*\\
			CFNet2-tri \cite{Siam-tri}&78.0&59.2&55.3*\\
			RFL \cite{staple}&77.8&58.1&15.0*\\
			SiamFC \cite{siamesefc}&77.1&58.2&86.0*\\
			DLSSVM \cite{dlssvm}&76.3&53.9&4.4*\\
			CFNet \cite{cfnet}&74.8&56.8&75.0*\\
			KCF \cite{kcf}&69.6&47.7&{\color{red}170.4}\\
			CNT \cite{cnt}&57.2&45.2&1.5\\
			\bottomrule
		\end{tabular}
	}
\end{table}

\cref{table_otb} compares the precision scores, AUC scores and speeds obtained by our OA-LSTM-ADA and other state-of-the-art trackers. For the tracking speed, KCF is the fastest among the compared trackers, but it achieves the worse tracking accuracy than other recent state-of-the-art trackers. SiamFC, CFNet,  SiamFC-tri, CFNet2-tri and MemTrack achieve high speeds and competitive tracking accuracy owing to the efficiency of the Siamese network. But they are worse than our OA-LSTM-ADA for both the precision and AUC scores. Our OA-LSTM-ADA performs better than high-speed KCF and TRACA (with speeds beyond 100 fps) in tracking accuracy while still maintaining a real-time speed. Staple, CSR-DCF and CNT are able to operate at satisfactory speeds on CPU. However, their tracking accuracies are much lower than our OA-LSTM-ADA. Other trackers, \emph{i.e.},  CNN-SVM, ACFN, RFL and DLSSVM, are slower and less accurate than our OA-LSTM-ADA. These results demonstrate that OA-LSTM-ADA achieves outstanding trade-off in terms of state-of-the-art accuracy and real-time speed among all the competing trackers.

\subsubsection{Attribute-Based Comparison}

\begin{figure*}[!tb]
	\centering
	\hspace*{-2em}
	\includegraphics[height=0.245\textwidth]{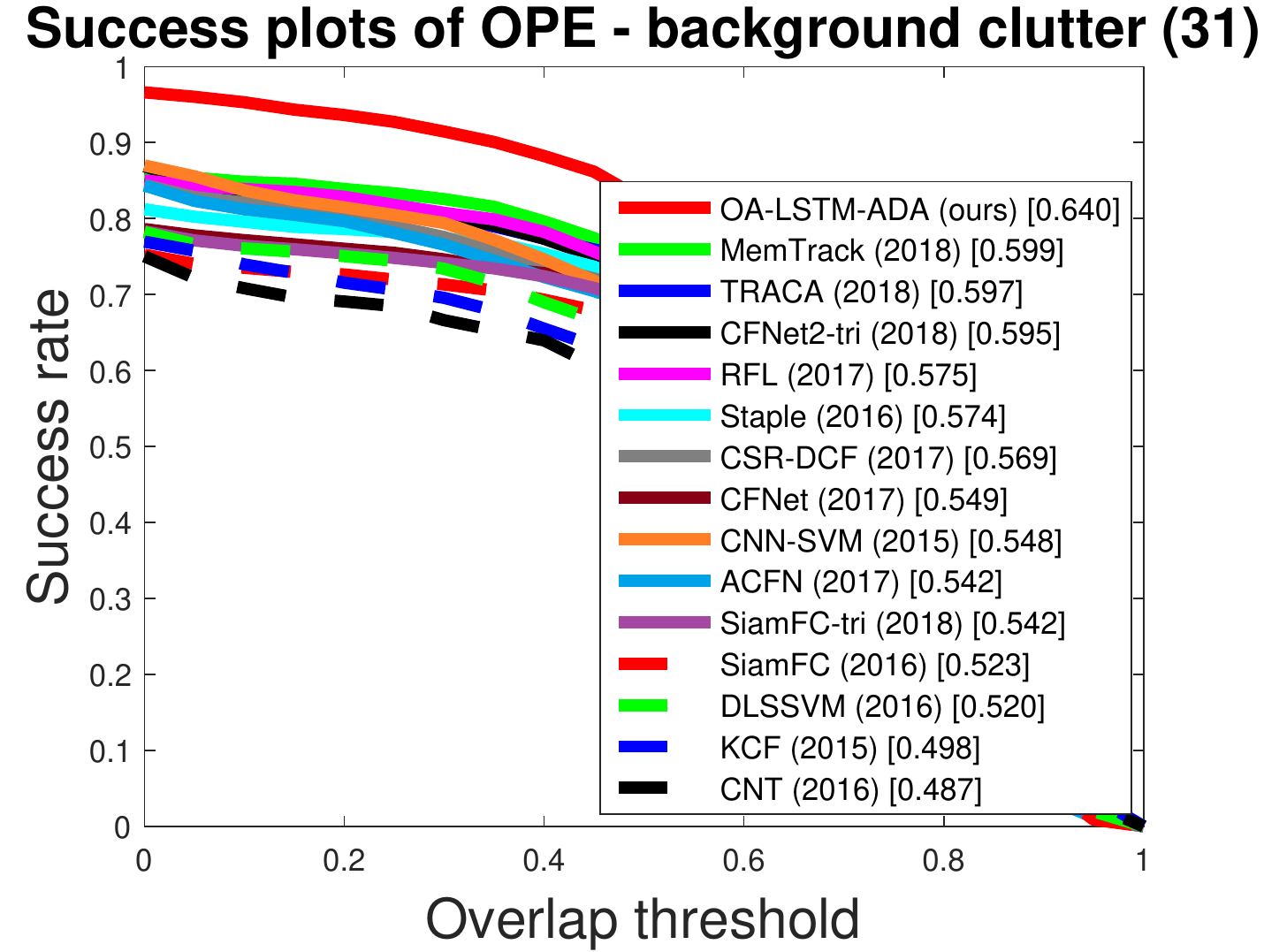}
	\includegraphics[height=0.25\textwidth]{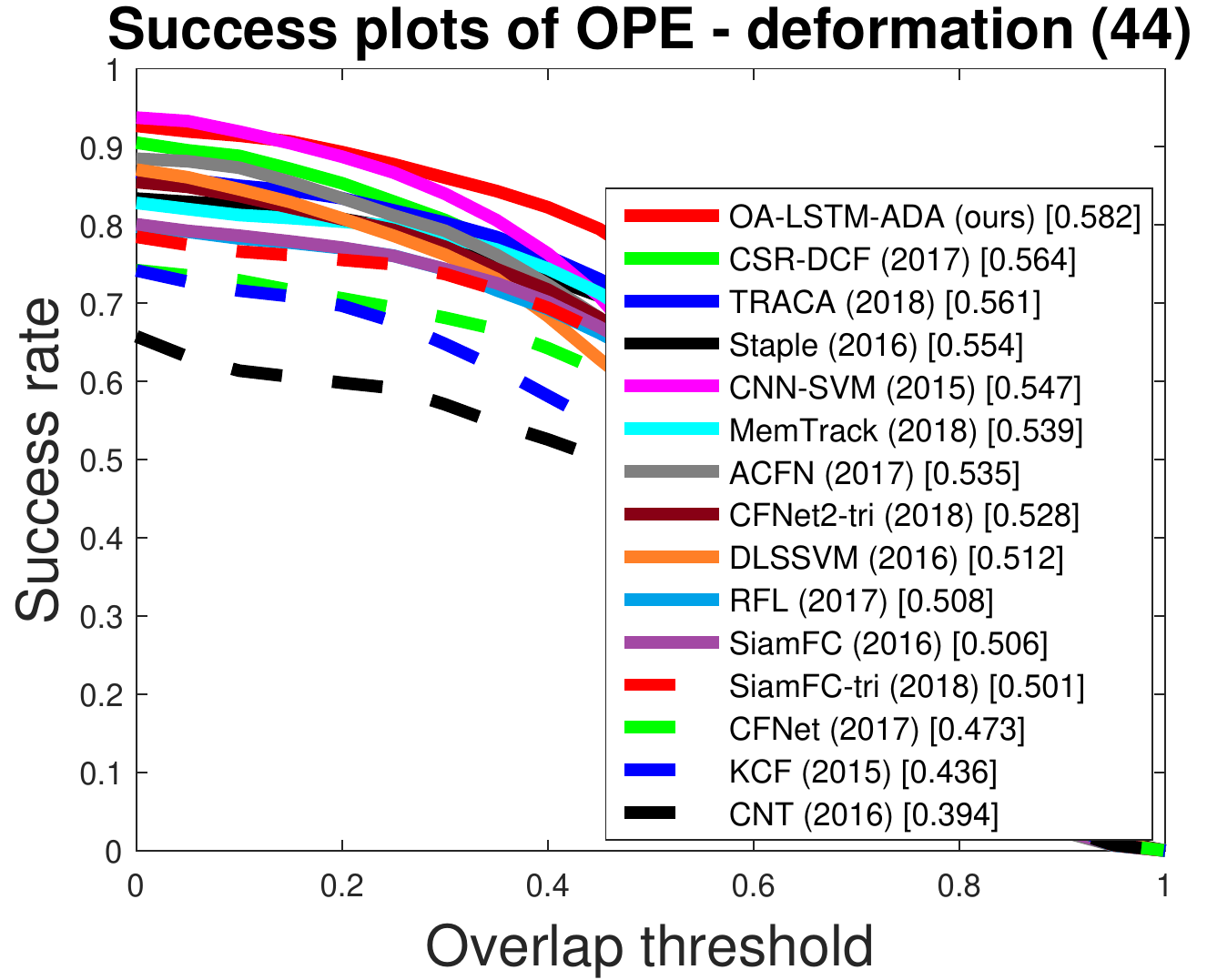}
	\includegraphics[height=0.25\textwidth]{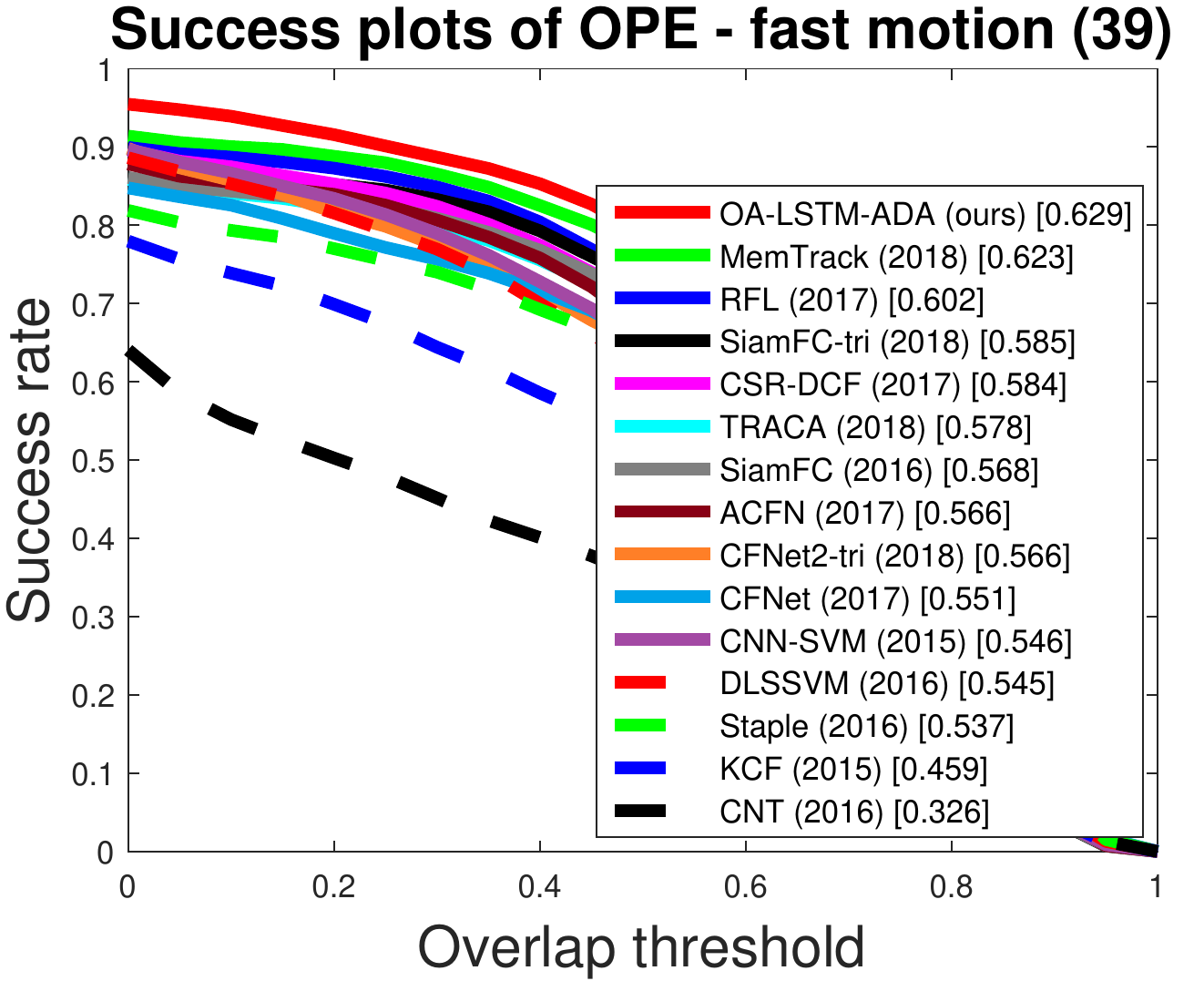}
	\includegraphics[height=0.25\textwidth]{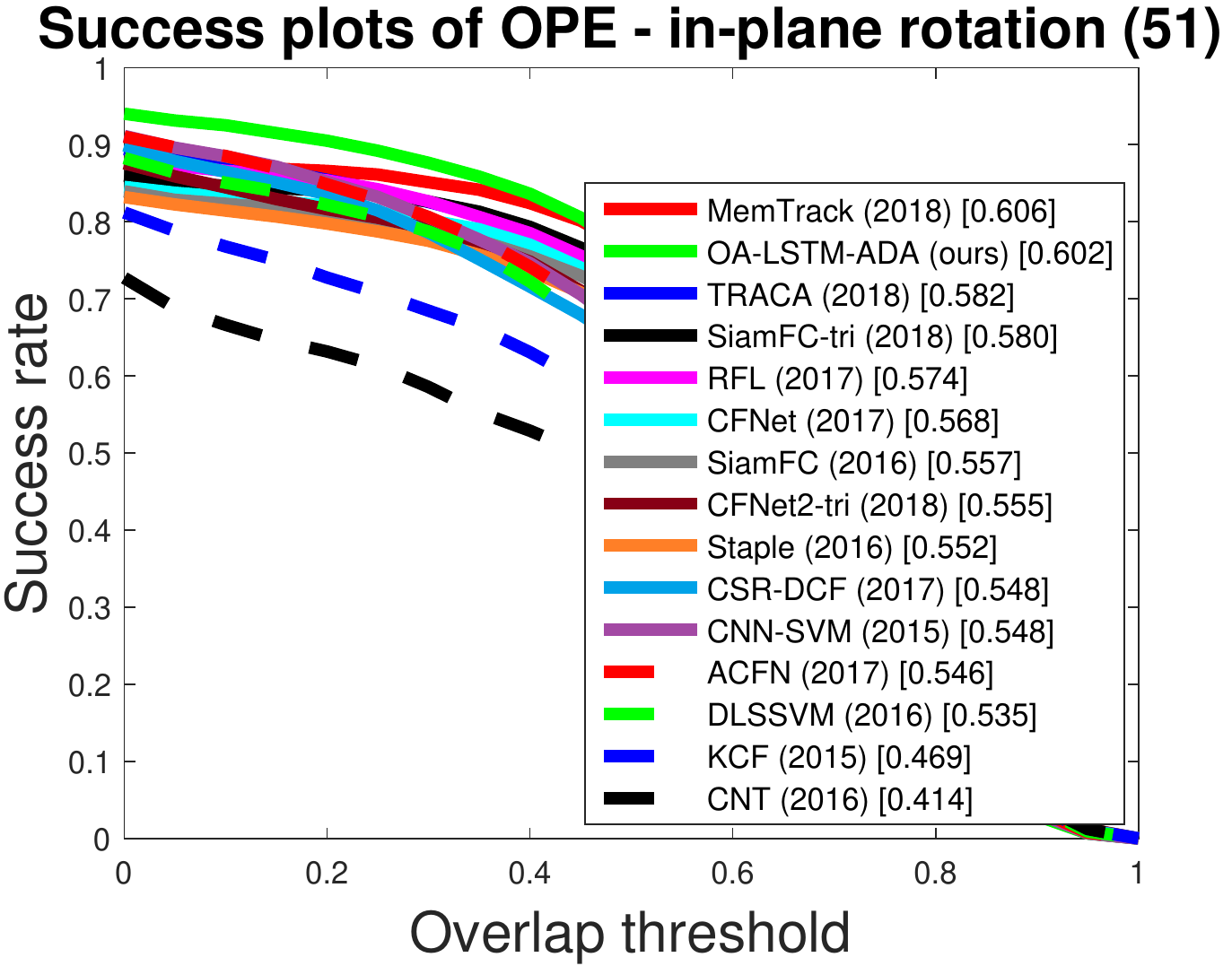}
	\includegraphics[height=0.25\textwidth]{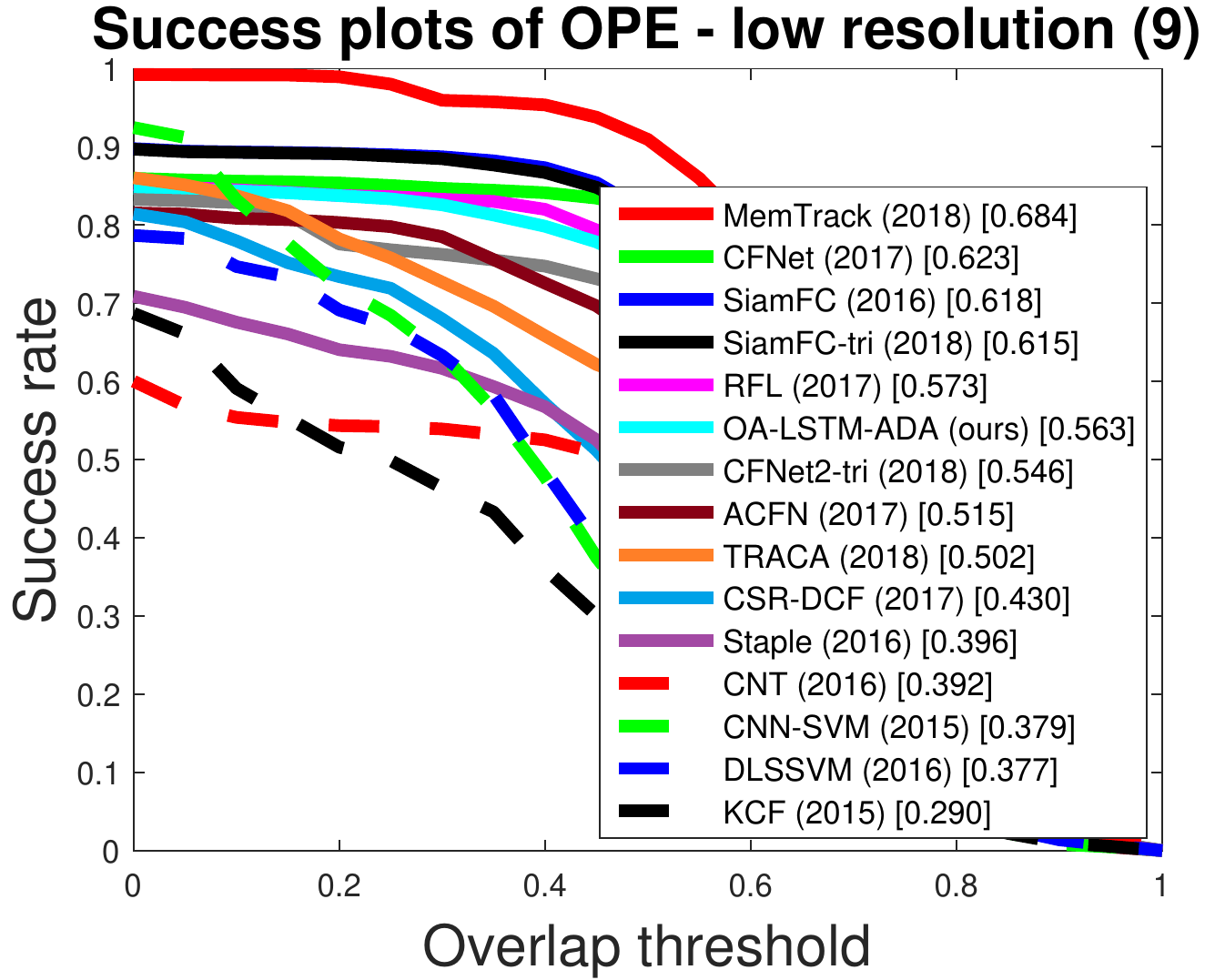}
	\includegraphics[height=0.25\textwidth]{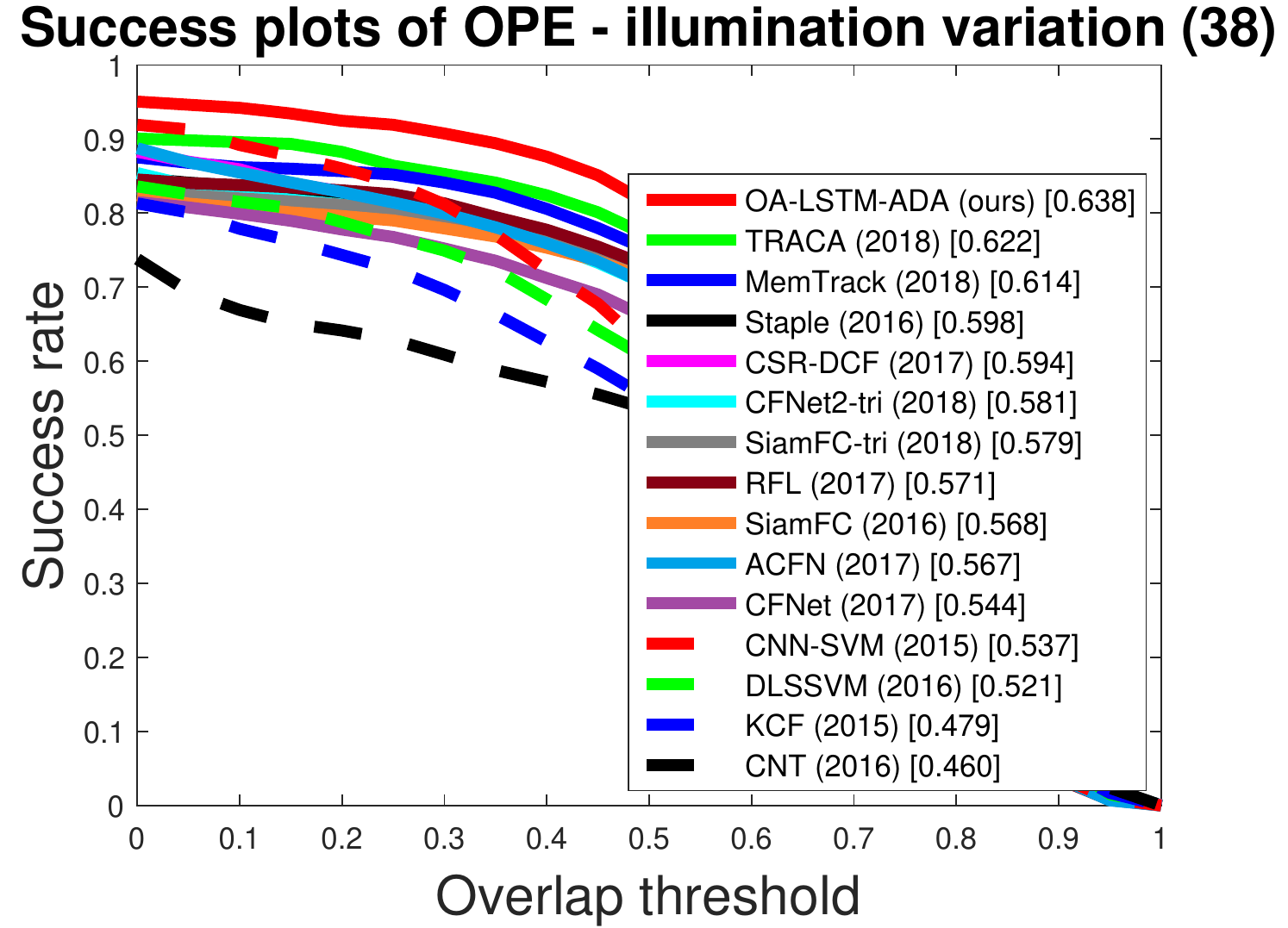}
	\includegraphics[height=0.25\textwidth]{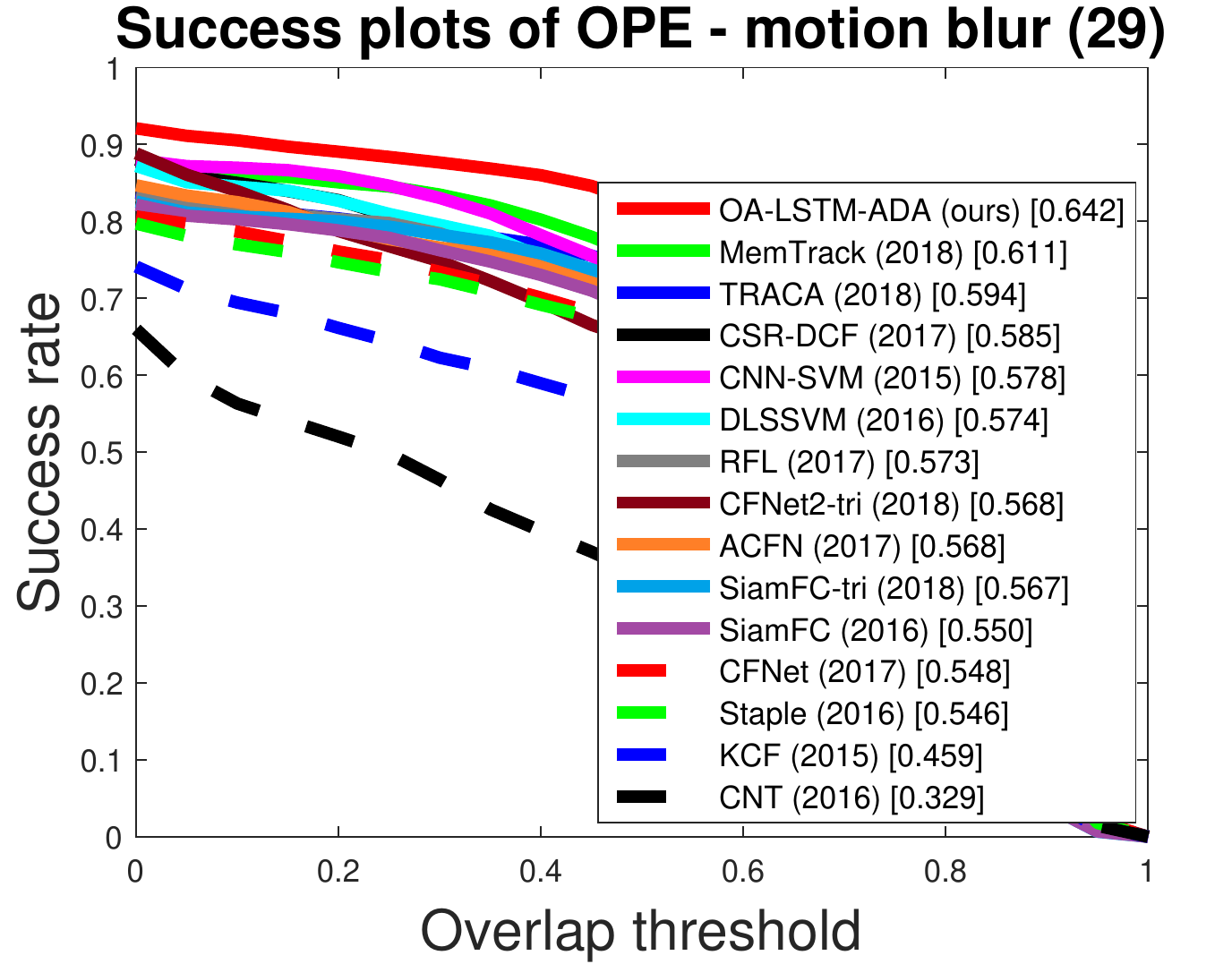}
	\includegraphics[height=0.25\textwidth]{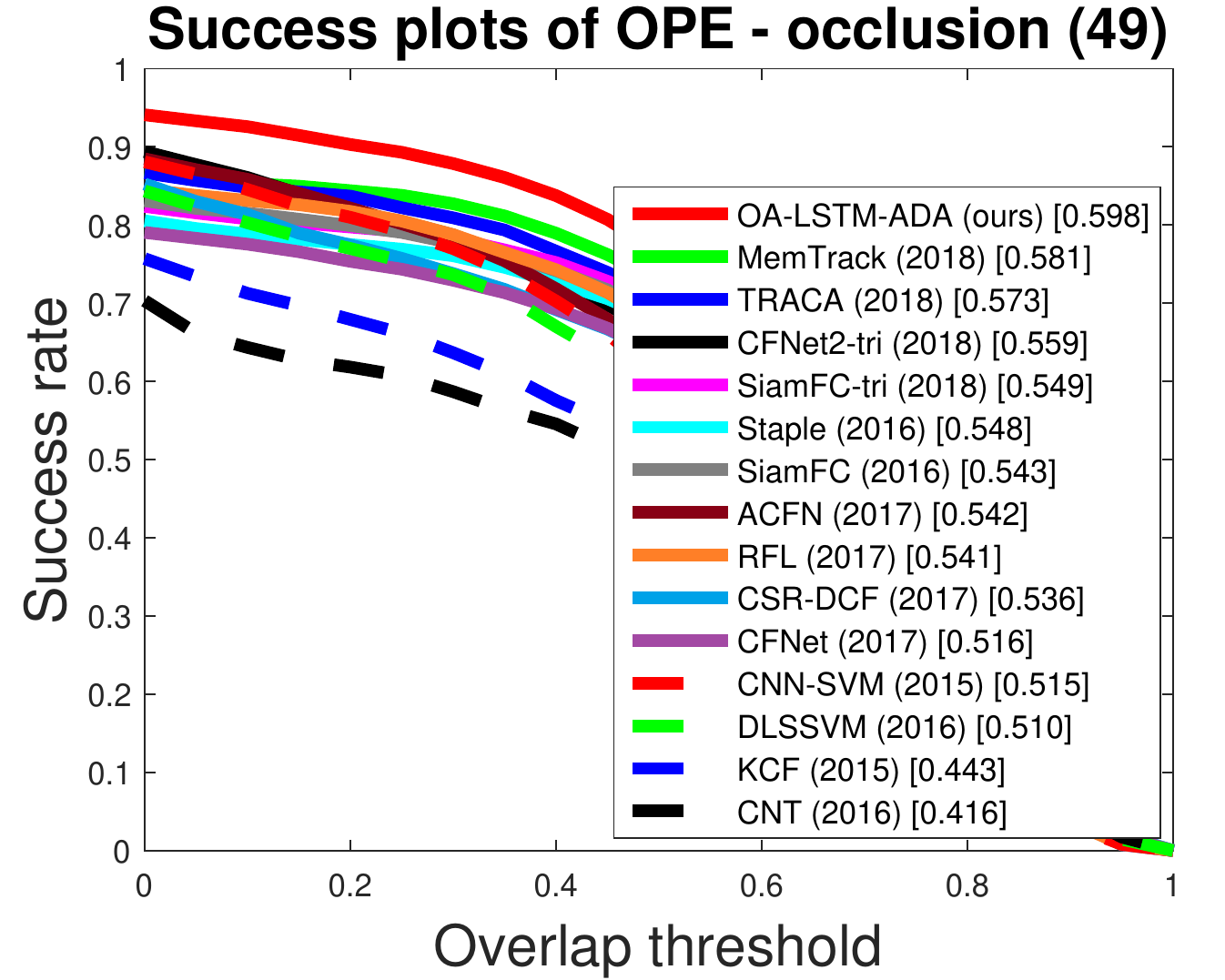}
	\includegraphics[height=0.25\textwidth]{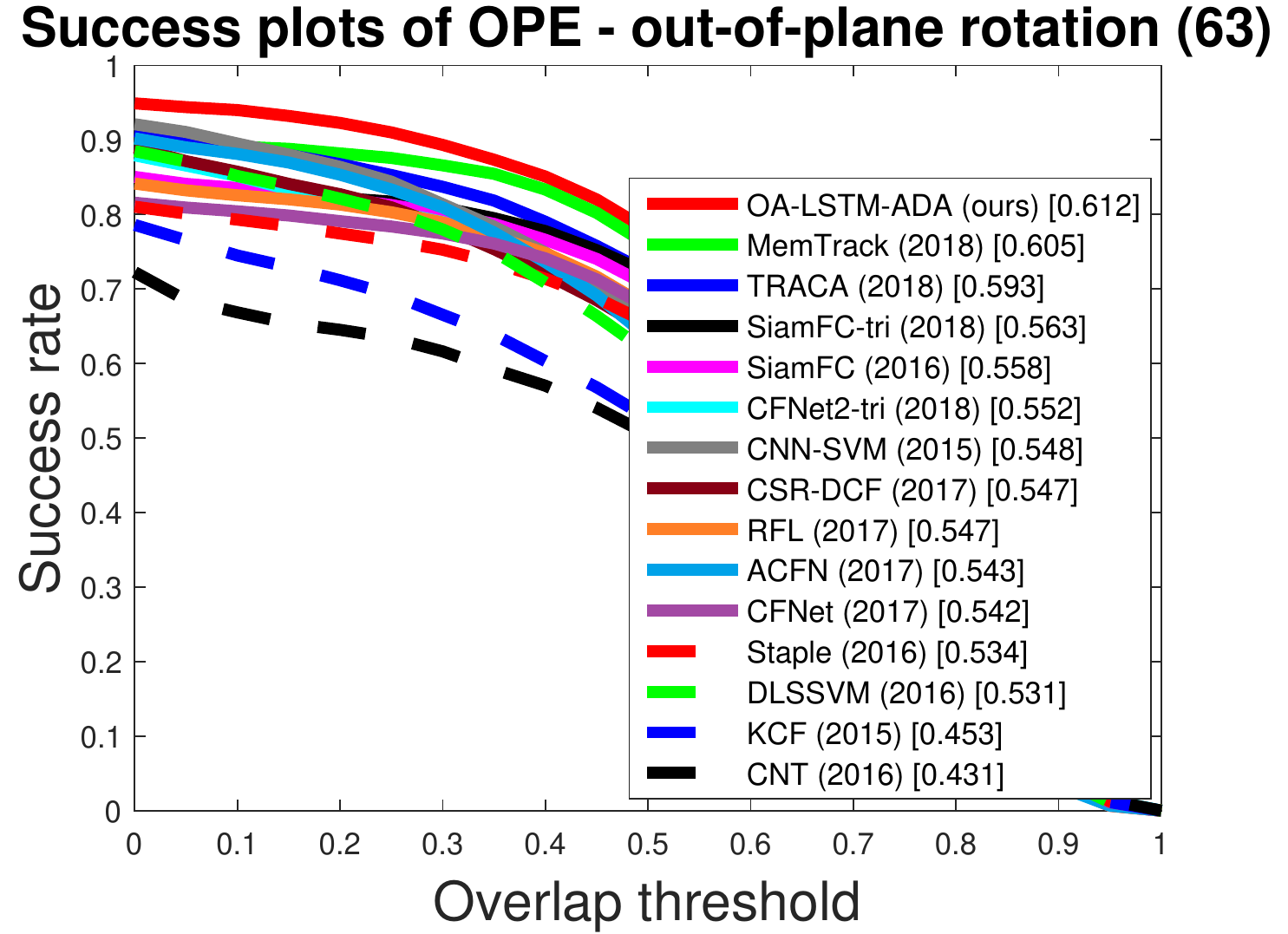}
	\includegraphics[height=0.25\textwidth]{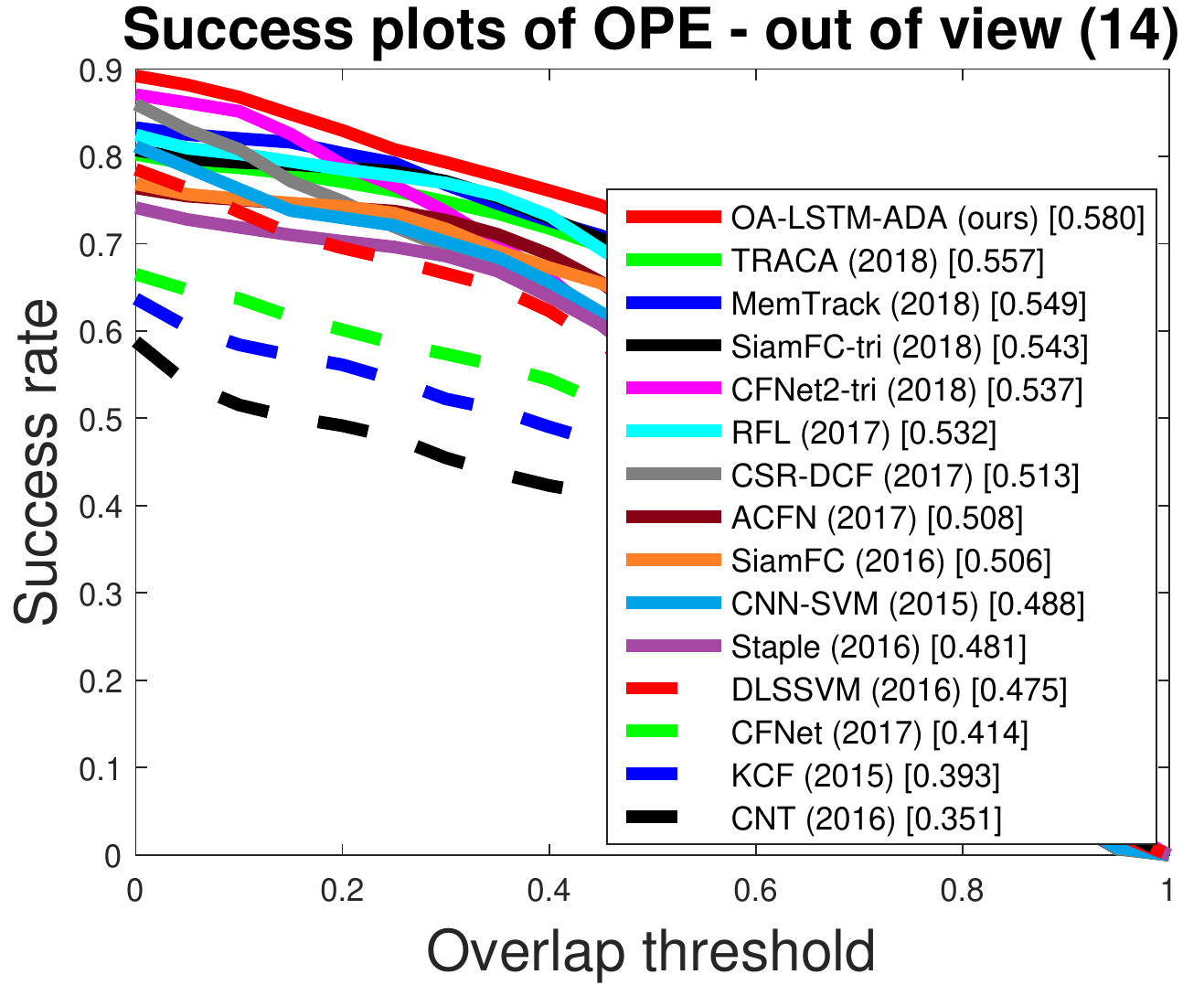}
	\includegraphics[height=0.25\textwidth]{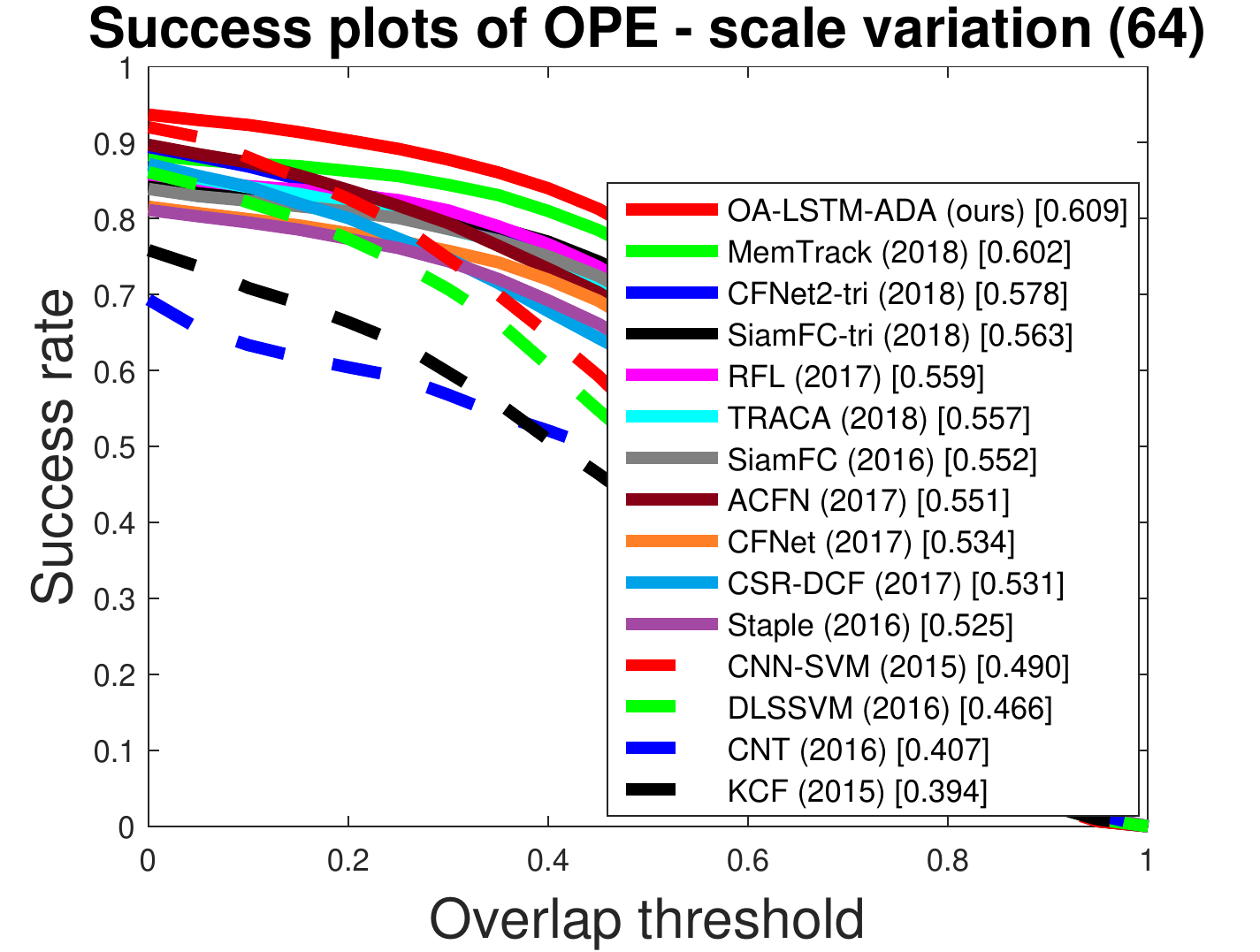}
	
	\caption{The success plots on the OTB-2015 dataset for eleven challenging attributes: background clutter, deformation, fast motion, in-plane rotation, low resolution, illumination variation, motion blur, occlusion, out-of-plane rotation, out of view and scale variation. }
	\label{fig_otb_attribute}
\end{figure*}

\cref{fig_otb_attribute} compares the performance obtained by our OA-LSTM-ADA and other state-of-the-art trackers using success plots on the OTB-2015 dataset for eleven challenging attributes including  background clutter, deformation, fast motion, in-plane rotation, low resolution, illumination variation, motion blur, occlusion, out-of-plane rotation, out of view and scale variation.

Our OA-LSTM-ADA performs favorably against other compared state-of-the-art trackers in most cases, which indicates that OA-LSTM-ADA possesses high robustness while operating in real-time. Compared with the representative Siamese network based tracker, \emph{i.e.}, SiamFC, our OA-LSTM-ADA achieves significant performance improvements under all the eleven challenge attributes. This clearly proves that the proposed object-adaptive LSTM network is able to effectively utilize the sequential dependencies among successive frames and learn the object appearance variations with high online adaptability. OA-LSTM-ADA outperforms the reccurent trackers, \emph{i.e.}, MemTrack and RFL, under most attributes, which demonstrates the robustness of our LSTM network for classification, compared with the LSTM networks for object template management used in MemTrack and RFL. OA-LSTM-ADA obtains much better performance than other compared trackers in the presence of fast motion, occlusion and out of view. This is because that OA-LSTM-ADA can memorize the previous object appearance and ignore the distracting similar objects via the object-adaptive LSTM network. For the attributes of in-plain rotation and low resolution, OA-LSTM-ADA performs  worse than MemTrack. The reason may be that the object template used for similarity computing lacks effective updating and thus deviates from the temporal object under such  disturbances at the later stage of tracking. Even so, OA-LSTM-ADA obtains a higher tracking accuracy than MemTrack on the whole dataset.

\subsubsection{Qualitative Comparison}
\begin{figure*}[!tb]
	\centering
	\includegraphics[width=1\textwidth]{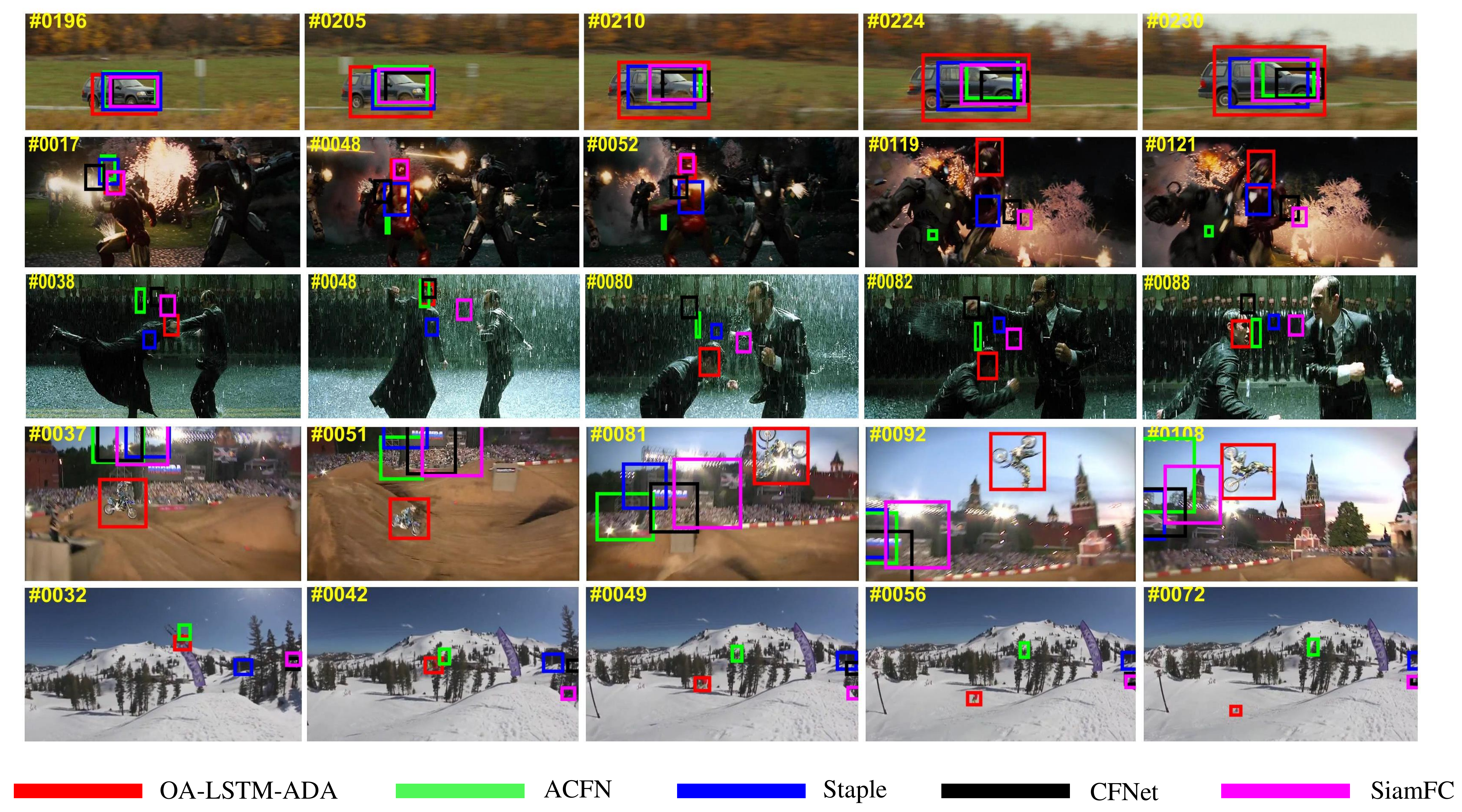}
	\caption{Qualitative results of our OA-LSTM-ADA, ACFN \cite{acfn}, Staple \cite{staple}, CFNet \cite{cfnet} and SiamFC \cite{siamesefc} on five challenging sequences (from top to down: \emph{CarScale}, \emph{Ironman}, \emph{Matrix}, \emph{MotorRolling} and \emph{Skiing}, respectively).}
	\label{fig_otb_qualitative}
\end{figure*}

\cref{fig_otb_qualitative} qualitatively compares the performance obtained by our OA-LSTM-ADA, ACFN, Staple, CFNet and SiamFC on five challenging sequences.

For the most challenging sequences, most trackers fail to locate the target position or incorrectly estimate the target scale, while our OA-LSTM-ADA accurately tracks the object in terms of both position and scale. For the sequence of \emph{CarScale} (row 1), the compared trackers are able to correctly locate the target position, but they only discriminate a part of the object instead of the whole object when the object undergoes large scale variation. In spite of the challenging scale variation, our OA-LSTM-ADA correctly estimates both the position and scale of the object. For the sequences of \emph{Ironman} and \emph{Matrix} (row 2 and row 3), the most compared trackers drift away because of the significant illumination variation and occlusion. In contrast, our OA-LSTM-ADA successfully handles these challenges and accurately tracks the object despite the complex backgrounds. In the sequences of \emph{MotorRolling} and \emph{Skiing} (row 4 and row 5), the compared trackers struggle when encountering fast motion and significant rotation, while our OA-LSTM-ADA keeps robust tracking of the object throughout the sequence.

\subsection{Evaluation on TC-128}
\label{section_tc128}
\subsubsection{Dataset and Evaluation Metrics}
The TC-128 \cite{tc128} dataset contains 128 fully annotated color video sequences with many challenging factors.
Similar to the evaluation on OTB (\cref{otb-metrics}), we also use the performance evaluation method of OPE and metrics of precision plots and success plots for the evaluation on TC-128.


\subsubsection{Quantitative Comparison}

\begin{figure*}[!tb]
	\centering
	\label{OTB-2015-pre}
	\includegraphics[width=0.49\textwidth]{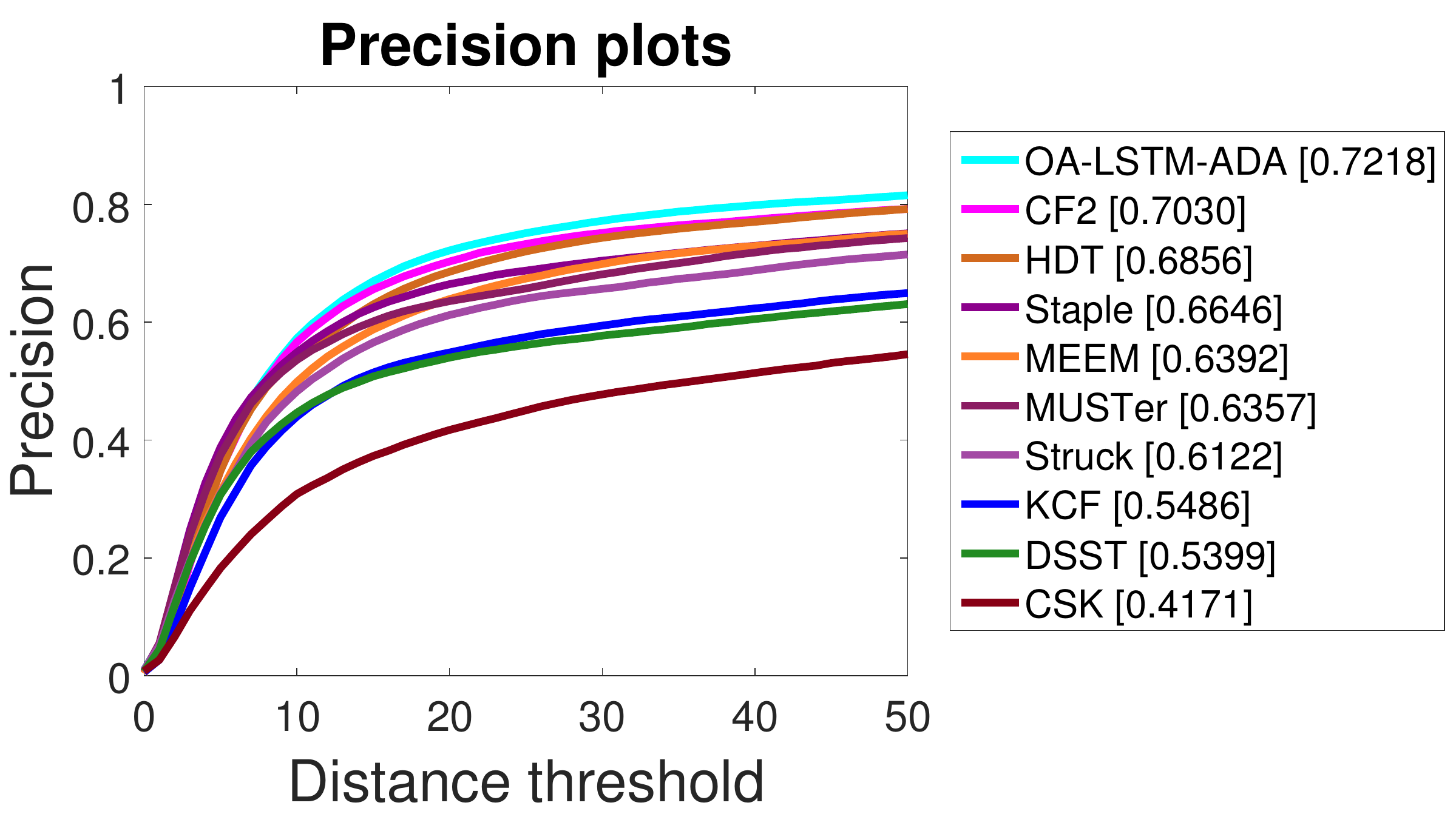}
	\label{OTB-2015-suc}
	\includegraphics[width=0.49\textwidth]{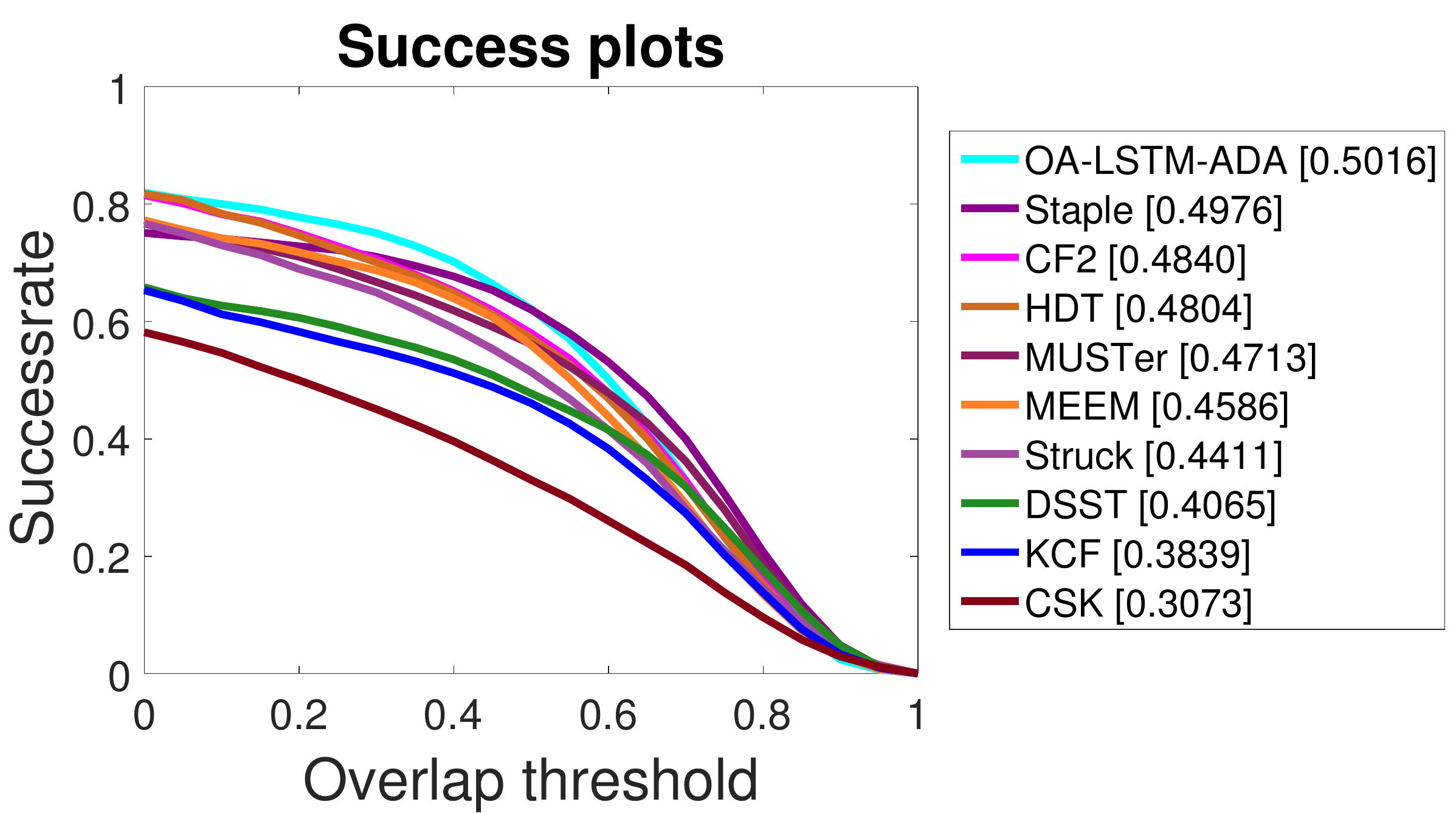}
	\caption{Precision plots and success plots showing the performance of our OA-LSTM-ADA compared with other state-of-the-art trackers on the TC-128 dataset.}
	\label{fig^tc-128}
\end{figure*}

\begin{table}[!tb]
	\centering
	\caption{The precision
		score, the AUC (Area Under the Curve) score and
		speed (fps, * indicates GPU speed, otherwise CPU speed) on the TC-128 dataset. The best and second best results are displayed in red and blue fonts, respectively.}
	\label{table^tc128}
	\setlength{\tabcolsep}{6mm}
	\renewcommand\arraystretch{0.8}
	\scalebox{0.8}{
	\begin{tabular}{cccc}
		\toprule
		Tracker&Precision&AUC&Speed\\
		\midrule
		\textbf{OA-LSTM-ADA}&{\color{red}72.18}&{\color{red}50.16}&32.5*\\
		CF2 \cite{hcf}&{\color{blue}70.30}&48.40&10.8\\
		HDT \cite{hdt}&68.56&48.04&9.7\\
		Staple \cite{staple}&66.46&{\color{blue}49.76}&50.8\\
		MEEM \cite{meem} &63.92&45.86&11.1\\
		MUSTer \cite{muster}&63.57&47.13&4.0\\
		Struck \cite{struck}&61.22&44.11&17.8\\
		KCF \cite{kcf}&54.86&38.39&{\color{blue}170.4}\\
		DSST \cite{dsst}&53.99&40.65&12.5\\
		CSK \cite{csk}&41.71&30.73&{\color{red}269.0}\\
		\bottomrule
	\end{tabular}
}
\end{table}

We quantitatively compare our OA-LSTM-ADA with several state-of-the-art trackers including CF2 \cite{hcf}, HDT \cite{hdt}, Staple \cite{staple}, MEEM \cite{meem}, MUSTer \cite{muster}, Struck \cite{struck}, KCF \cite{kcf}, DSST \cite{dsst} and CSK \cite{csk}. \cref{fig^tc-128} shows the comparative results in terms of precision plots and success plots on the TC-128 \cite{tc128} dataset.

We can observe that our OA-LSTM-ADA achieves the best performance in both precision plots and success plots among all the compared trackers. OA-LSTM-ADA outperforms the other two trackers which also use deep features, \emph{i.e.}, CF2 and HDT, with relative improvements of 1.88\% (1.76 \%) and 3.62\% (2.12\%), respectively. Compared with the trackers based on the hand-crafted features, such as Staple and MEEM, our OA-LSTM-ADA achieves higher tracking accuracy and obtains a real-time speed on the GPU.

\cref{table^tc128} presents the precision scores, AUC scores and speeds obtained by our OA-LSTM-ADA and other compared state-of-the-art trackers.

As shown in \cref{table^tc128}, our OA-LSTM-ADA performs favorably against other state-of-the-art trackers in terms of both precision scores and AUC scores while maintaining a real-time speed. Compared with fast correlation filter based trackers such as KCF \cite{kcf} and Staple \cite{srdcf}, which can operate at high speeds on a CPU, our OA-LSTM-ADA achieves noticeably accuracy improvements in both precision scores and AUC scores. Compared with the correlation filter based trackers using deep features, such as CF2 and HDT,  our OA-LSTM-ADA shows the performance superiority. This indicates that the proposed object-adaptive LSTM network can effectively adapt to the temporarily changing object and is well suited for the visual tracking task. In addition, the proposed fast proposal selection strategy provides high efficiency for our deep model, which allows our tracker to be performed at real-time speed. MEEM exploits a multi-expert restoration scheme to handle the drift problem during online tracking. MUSTer adopts cognitive psychology principles to design an adaptive representation for visual tracking. Although these trackers can be performed on a CPU, there still exists a gap between their tracking accuracy and that of our OA-LSTM-ADA.

\subsection{Evaluation on UAV-123}
\label{section_uav123}
\subsubsection{Dataset and Evaluation Metrics}
The UAV-123 \cite{uav123} dataset consists of 123 fully annotated video sequences captured from a low-altitude aerial perspective  for  UAV target tracking. Similar to the evaluations on OTB in \cref{section_otb} and TC-128 in \cref{section_tc128}, we use the OPE performance evaluation method and metrics of precision plots and success plots to conduct the experiments on UAV-123.

\subsubsection{Quantitative Comparison}

\begin{figure*}[!tb]
	\centering
	\includegraphics[width=0.48\textwidth]{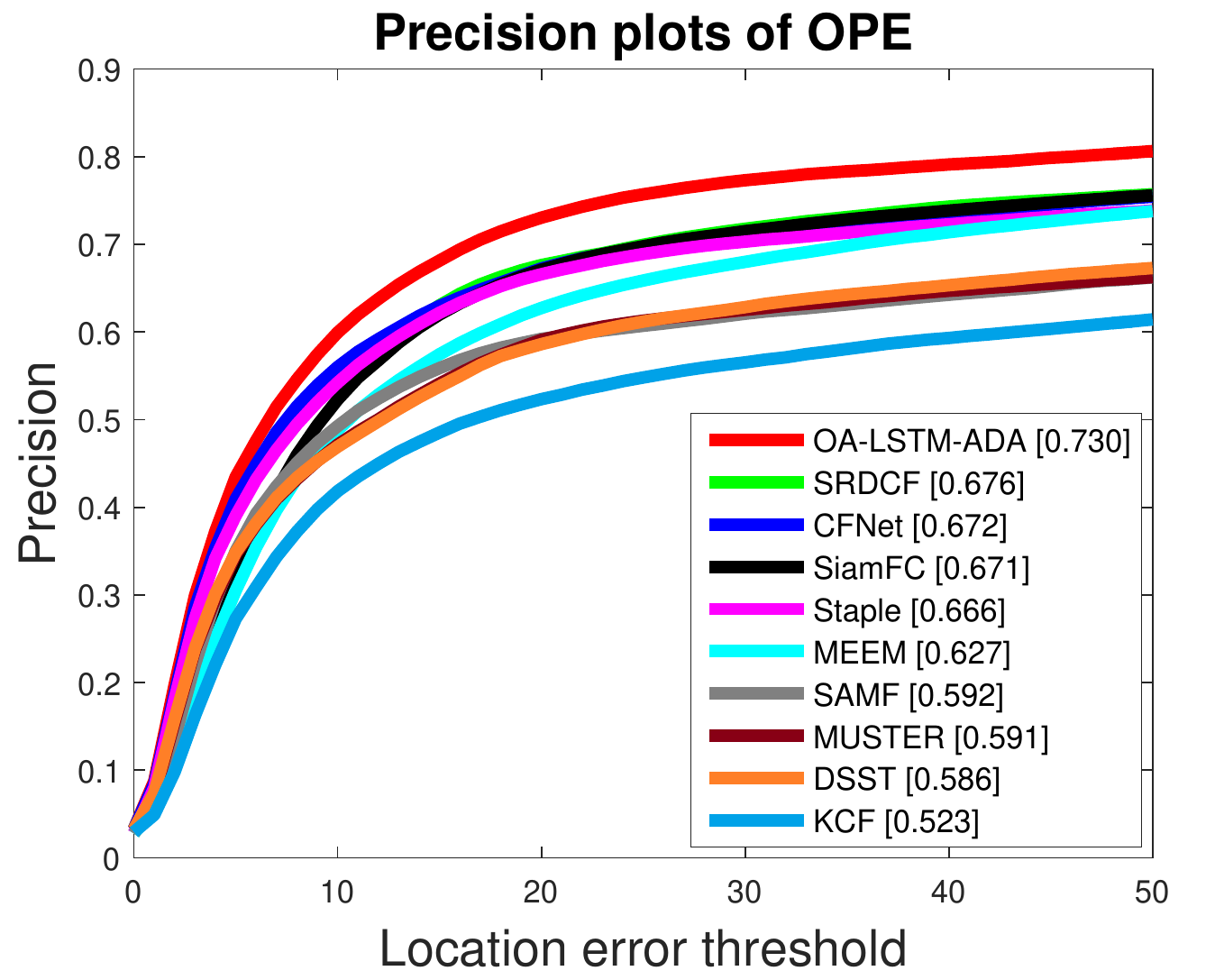}
	\includegraphics[width=0.48\textwidth]{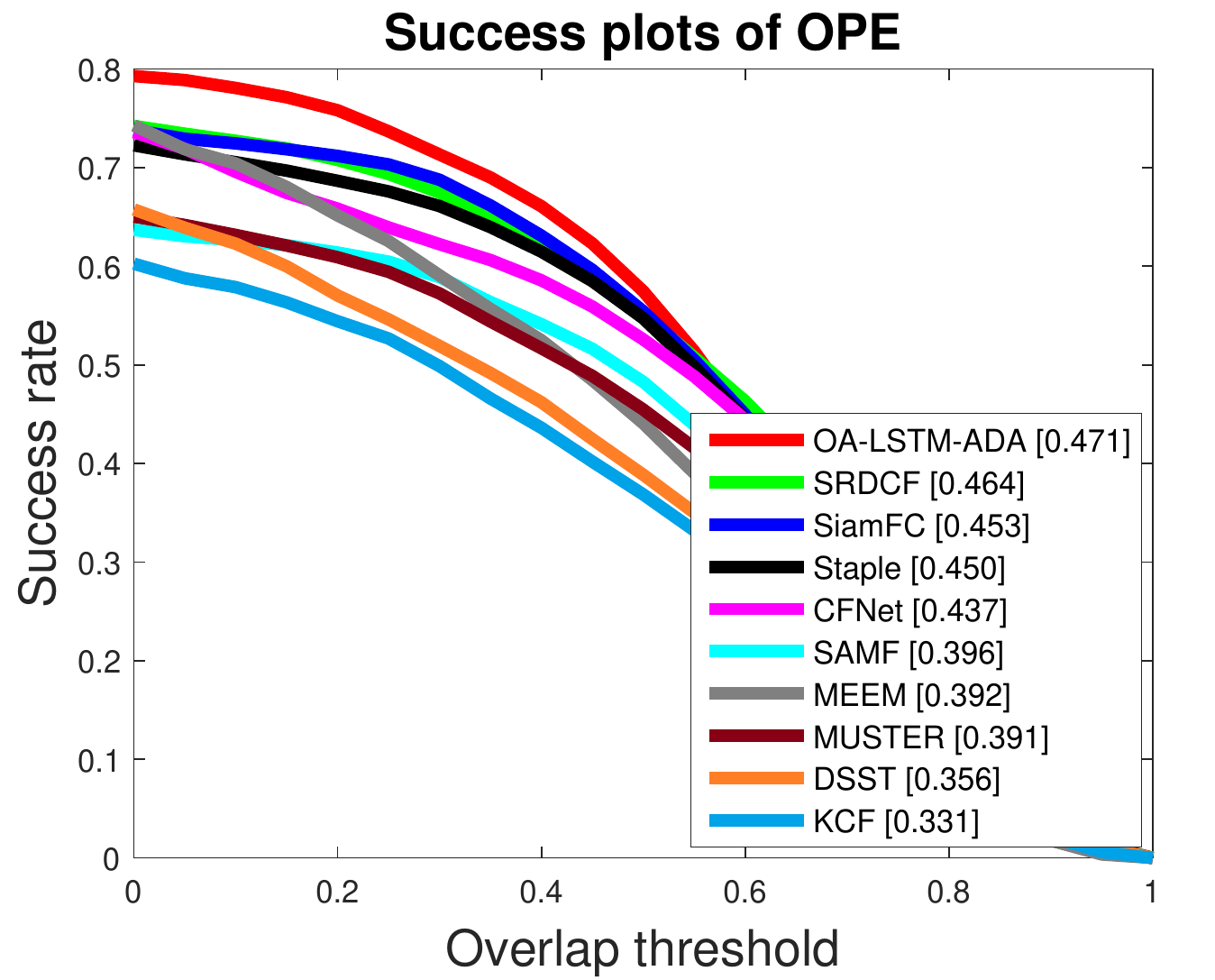}
	\caption{Precision plots and success plots showing the performance of our OA-LSTM-ADA compared with other state-of-the-art trackers on the UAV-123 dataset.}
	\label{fig_UAV-123}
\end{figure*}

\cref{fig_UAV-123} shows the quantitative comparison of our OA-LSTM-ADA and several state-of-the-art trackers that have publicly available results on the UAV-123 dataset, including SRDCF \cite{srdcf}, CFNet \cite{cfnet}, SiamFC \cite{siamesefc}, Staple \cite{staple}, MEEM \cite{meem}, SAMF \cite{samf}, MUSTER \cite{muster}, DSST \cite{dsst} and  KCF \cite{kcf}.
In terms of both precision and success plots, our OA-LSTM-ADA outperforms all the other trackers with a real-time speed. Compared with the Siamese network based trackers, \emph{i.e.}, SiamFC \cite{siamesefc} and CFNet \cite{cfnet}, our OA-LSTM-ADA achieves a higher tracking accuracy owing to the effectiveness of the proposed object-adaptive LSTM network and data augmentation technique. Compared with the hand-crafted feature based trackers, such as SRDCF \cite{srdcf} and Staple \cite{staple}, our OA-LSTM-ADA, which uses deep features and adopts an efficient object-adaptive LSTM network with fast proposal selection, achieves better performance while maintaining a real-time speed.

\subsection{Evaluation on VOT-2017}
\label{section_vot}
\subsubsection{Dataset and Evaluation Metrics}

The VOT-2017 \cite{vot2017} dataset contains 60 fully annotated video sequences. The performance evaluation metric is the Expected Average Overlap (EAO) score, which takes both accuracy and robustness into account. The speed is reported in terms of EFO, which normalizes speed measurements obtained over different hardware platforms. VOT-2017 introduces a new real-time challenge, where trackers are required to deal with the video frames at real-time speeds.  We evaluate the proposed method  on the VOT-2017 real-time challenge.

\subsubsection{Quantitative Comparison}

\begin{figure*}[!tb]
	\centering
	\hspace*{-1em}
	\includegraphics[width=0.9\textwidth]{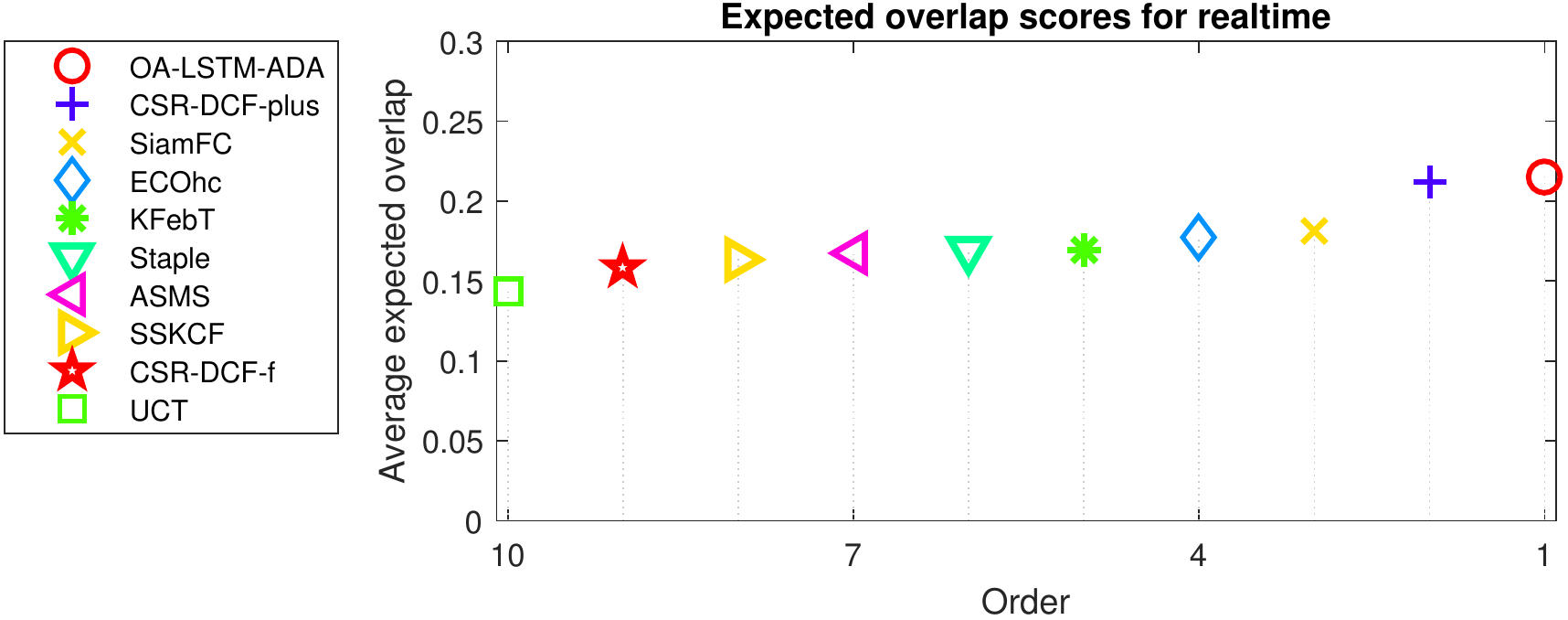}
	\caption{Expected Average Overlap (EAO) ranking on the VOT-2017  real-time  challenge. We compare our OA-LSTM-ADA with the top 9 trackers on this challenge.}
	\label{fig_VOT-2017}
\end{figure*}

\begin{table}[!tb]
	\centering
	\caption{The Expected Average Overlap (EAO) score and
		speed (in EFO units) on the VOT-2017  real-time challenge. The best and second best results are displayed in red and blue fonts, respectively.}
	\setlength{\tabcolsep}{1.5mm}
	\renewcommand\arraystretch{0.8}
	\scalebox{0.8}{
		\label{table_vot2017}
		\begin{tabular}{cccccccccccc}
			\toprule
			Tracker & \textbf{OA-LSTM-ADA} & CSR-DCF-plus \cite{csr-dcf}  & SiamFC \cite{siamesefc} & ECOhc  \cite{ECO} & KFebT   \cite{kfebt}
			\\
			\midrule
			EAO & {\color{red}0.216} & {\color{blue}0.212} & 0.182 & 0.177 & 0.170
			\\
			EFO & 3.12 & 4.59 & 5.33 & 4.69 & {\color{blue}30.22}
			\\
			\midrule
			Tracker & Staple \cite{staple} & ASMS \cite{asms} & SSKCF \cite{sskcf} & CSR-DCF-f \cite{csr-dcf} & UCT  \cite{uct}
			\\
			\midrule
			EAO & 0.169 & 0.167 & 0.164 & 0.158 & 0.144
			\\
			EFO & 8.19 & {\color{red}34.03} & 7.99 & 2.88 & 3.09
			\\
			\bottomrule
		\end{tabular}
	}
\end{table}


We compare our OA-LSTM-ADA with the top 9 trackers on the VOT-2017 real-time challenge, including CSR-DCF-plus \cite{csr-dcf}, CSR-DCF-f \cite{csr-dcf}, SiamFC \cite{siamesefc}, ECOhc \cite{ECO}, Staple \cite{staple}, KFebT \cite{kfebt}, ASMS \cite{asms}, SSKCF and  UCT \cite{uct}.
\cref{fig_VOT-2017} presents the Expected Average Overlap (EAO) ranking on the VOT-2017  real-time  challenge. \cref{table_vot2017} illustrates specific EAO scores and speeds  (in EFO units)  of the compared trackers. Our OA-LSTM-ADA ranks first with the EAO score of 0.216 in this challenge, while maintaining a real-time speed. In particular, OA-LSTM-ADA shows a significant improvement over its baseline SiamFC, which verifies the effectiveness and efficiency of the proposed object-adaptive LSTM network and data augmentation technique.

\section{Conclusions and Future Work}
\label{section_conclusion}
In this paper, we propose a novel object-adaptive LSTM network for real-time tracking, which can effectively capture temporal dependencies in the video sequence and dynamically adapt to the temporarily changing object. The LSTM network is learned online based on the  sequence-specific information. Thus, it is able to robustly track an arbitrary object without the risk of over-fitting to the tracking datasets. In order to improve the computational efficiency, we also propose a fast proposal selection strategy. This strategy utilizes the matching-based tracking method to pre-estimate the dense proposals and select high-quality ones to feed to the LSTM network for further evaluation. In this way, the computational burden rendered by the irrelevant proposals is alleviated so that the proposed method can operate in real-time. Moreover, to handle the problems of sample inadequacy and class imbalance during the online learning of the LSTM network, we also use GAN to augment the available training data. This data augmentation technique facilitates the training of the LSTM network and improves the tracking performance. Extensive experiments on the OTB \cite{otb15}, TC-128 \cite{tc128}, UAV-123 \cite{uav123} and VOT-2017 \cite{vot2017} benchmarks demonstrate the superior performance of the proposed method at  the real-time speed compared with several state-of-the-art trackers. This exhibits great potentials of recurrent structures for visual tracking.

Future work will be directed towards  incorporating  attention prediction and aesthetics assessment  into our current tracking model, since such mechanisms may help to generate more high-quality proposals making full use of saliency information.  This can be achieved by designing a new attention-based recurrent network, and thus the performance of our tracking method may be further improved.

\section*{Acknowledgements}
	This work was supported by the National Key R\&D Program of China under Grant 2017YFB1302400, by the National Natural Science Foundation of China under Grants 61571379, U1605252 and 61872307, and by the Natural Science Foundation of Fujian Province of China under Grants 2017J01127 and 2018J01576.

\end{document}